\documentclass[10pt,twocolumn,letterpaper]{article}

\usepackage{iccv}
\usepackage{times}
\usepackage{epsfig}
\usepackage{graphicx}
\usepackage{amsmath}
\usepackage{amssymb}

\usepackage{multirow, subcaption, gensymb, float, slashbox}
\usepackage[dvipsnames]{xcolor}
\usepackage[breaklinks=true,bookmarks=false]{hyperref}

\iccvfinalcopy % *** Uncomment this line for the final submission

\def\iccvPaperID{3145} % *** Enter the ICCV Paper ID here

\ificcvfinal\fi
\begin{document}

%%%%%%%%% TITLE
\title{Adaptive Wing Loss for Robust Face Alignment via Heatmap Regression}

\author{Xinyao Wang$^{1,2}$ \qquad Liefeng Bo$^{2}$ \qquad Li Fuxin$^{1}$\\
$^{1}$Oregon State University \qquad\qquad $^{2}$JD Digits\\
{\tt\small \{wangxiny, lif\}@oregonstate.edu}, {\tt\small \{xinyao.wang3, liefeng.bo\}@jd.com}
}

\maketitle
\ificcvfinal\thispagestyle{empty}\fi

%%%%%%%%% ABSTRACT
\begin{abstract}
   Heatmap regression with a deep network has become one of the mainstream approaches to localize facial landmarks. However, the loss function for heatmap regression is rarely studied. In this paper, we analyze the ideal loss function properties for heatmap regression in face alignment problems. Then we propose a novel loss function, named Adaptive Wing loss, that is able to adapt its shape to different types of ground truth heatmap pixels. This adaptability penalizes loss more on foreground pixels while less on background pixels. To address the imbalance between foreground and background pixels, we also propose Weighted Loss Map, which assigns high weights on foreground and difficult background pixels to help training process focus more on pixels that are crucial to landmark localization. To further improve face alignment accuracy, we introduce boundary prediction and CoordConv with boundary coordinates. Extensive experiments on different benchmarks, including COFW, 300W and WFLW, show our approach outperforms the state-of-the-art by a significant margin on various evaluation metrics. Besides, the Adaptive Wing loss also helps other heatmap regression tasks. Code will be made publicly available at \url{https://github.com/protossw512/AdaptiveWingLoss}.
   
\end{abstract}
%%%%%%%%% BODY TEXT
\section{Introduction}
\label{sec:introduction}
\textbf{Face alignment}, also known as facial landmark localization, seeks to localize pre-defined landmarks on human faces. Face alignment plays an essential role in many face related applications such as face recognition~\cite{face_rec_1, face_rec_2, face_rec_3, face_rec_4, face_rec_5}, face frontalization~\cite{face_frontal_1, face_frontal_2, face_frontal_3} and 3D face reconstruction~\cite{face_recons_1, face_recons_2, face_recons_3, face_recons_4}. In recent years, Convolutional Neural Network (CNN) based heatmap regression has become one of the mainstream approaches for face alignment problems and achieved considerable performance on frontal faces. However, landmarks on faces with large pose, occlusion and significant blur are still challenging to localize.
\begin{figure}
    \centering
    \includegraphics[width=0.6\linewidth]{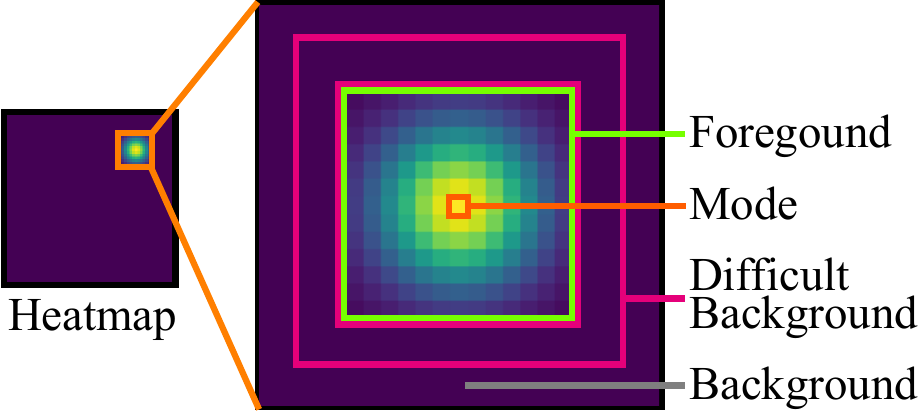}
    \vspace*{-0.1in}
    \caption{\textbf{Pixel type definitions.} (Best viewed in color).}
    \label{fig:pixel_definition}
    \vspace*{-0.1in}
\end{figure}
\begin{figure}[]
\captionsetup[subfigure]{justification=centering}
\begin{subfigure}{0.25\linewidth}
\centering
\includegraphics[width=0.9\linewidth]{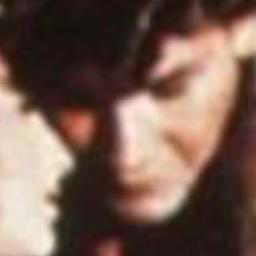}
\caption{Blurry \\ Large Pose}
\label{fig:ori_image}
\end{subfigure}%
\begin{subfigure}{0.25\linewidth}
\centering
\includegraphics[width=0.9\linewidth]{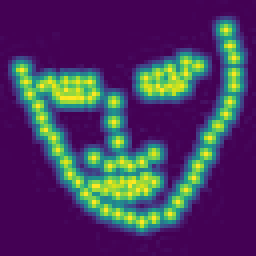}
\caption{GT \\ Heatmap}
\label{fig:gt_heatmap}
\end{subfigure}%
\begin{subfigure}{0.25\linewidth}
\centering
\includegraphics[width=0.9\linewidth]{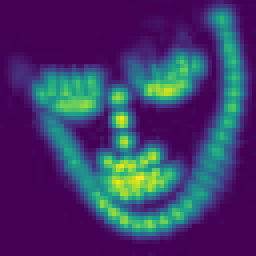}
\caption{MSE \\ NME: 6.27\%}
\label{fig:mse_heatmap}
\end{subfigure}%
\begin{subfigure}{0.25\linewidth}
\centering
\includegraphics[width=0.9\linewidth]{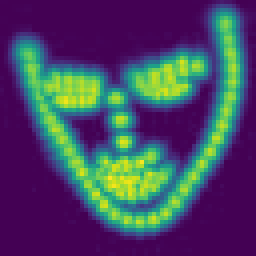}
\caption{AWing \\ NME: 4.23\%}
\label{fig:awing_heatmap}
\end{subfigure}%
\vspace*{-0.1in}
\caption{{\bf Predicted heatmap quality comparison}. The model trained with MSE failed to accurately predict the heatmap around left cheek, lower right cheek and eye brows. With the proposed Adaptive Wing loss (Fig.~\ref{fig:awing_heatmap}), the heatmap becomes much sharper on landmarks.}
\label{fig:Heatmap_comparsion}
\vspace*{-0.2in}
\end{figure}

\textbf{Heatmap regression}, which regresses a heatmap generated from landmark coordinates, is widely used for face alignment~\cite{heatmap_regre_1, heatmap_regre_2, heatmap_regre_3, heatmap_regre_4}. In heatmap regression, the ground truth heatmap is generated by plotting a Gaussian distribution centered at each landmark on each channel. The model regresses against the ground truth heatmap at pixel level and then use the predicted heatmaps to infer landmark locations. %See Figure~\ref{fig:pixel_definition} for pixel type definitions used in the rest of this paper.
Prediction accuracy on foreground pixels (pixels with positive values), especially the ones near the mode of each Gaussian distribution (Fig.~\ref{fig:pixel_definition}), is essential to accurately localize landmarks, even small prediction errors on these pixels can cause the prediction to shift from the correct modes. On the contrary, accurately predicting the values of background pixels (pixels with zero values) is less important, since small errors on these pixels will not affect landmark prediction in most cases. However, prediction accuracy on difficult background pixels (Fig.~\ref{fig:pixel_definition} background pixels near foreground pixels) are also important since they are often incorrectly regressed as foreground pixels and could cause inaccurate predictions.

From this discussion, we locate two issues of the widely used Mean Square Error (MSE) loss in heatmap regression: i) MSE is not sensitive to small errors, which hurts the capability to correctly locate the mode of the Gaussian distribution; ii) During training all pixels have the same loss function and equal weights, however, background pixels absolutely dominates foreground pixels on a heatmap. As a result of i) and ii), models trained with the MSE loss tend to predict a blurry and dilated heatmap with low intensity on foreground pixels compared to the ground truth (Fig.~\ref{fig:mse_heatmap}). This low quality heatmap could cause wrong estimation of facial landmarks. Wing loss~\cite{Feng_Wing_Loss} is shown to be effective to improve coordinate regression, however, according to our experiment, it is not applicable for heatmap regression. Small errors on background pixels will accumulate significant gradients and thus cause the training process to diverge. We thus propose a new loss function and name it Adaptive Wing loss (Sec.~\ref{sec:wing_loss}), that is able to significantly improve the quality of heatmap regression results.
\begin{figure*}
\centering
\vspace*{-0.1in}
\includegraphics[width=0.8\linewidth]{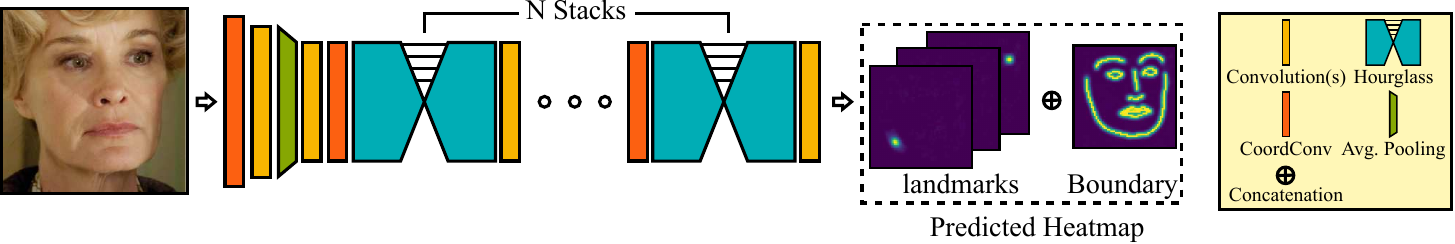}
\vspace*{-0.1in}
\caption{\textbf{An overview of our model}. The stacked HG takes a face image cropped with the ground truth bounding box and output one predicted heatmap for each landmark, respectively. An additional channel is used to predict facial boundaries. Due to limited space, we omitted the detailed structure of the stacked HG architecture, please refer ~\cite{Hourglass, bulat2017far} for details.}
\label{fig:model}
\vspace*{-0.2in}
\end{figure*}

Due to the translation invariance of the convolution operation in bottom-up and top-down CNN structures such as stacked Hourglass (HG)~\cite{Hourglass}, the network is not able to capture coordinate information, which we believe is useful for facial landmark localization, since the structure of human faces is relatively stable. Inspired by the CoordConv layer proposed by Liu \etal~\cite{CoordConv}, we encode into our model the full coordinate information and the information only on boundaries predicted from the previous HG module into our model. The encoded coordinate information further improves the performance of our approach. To encode boundary coordinates, we also add a sub-task of boundary prediction by concatenating an additional boundary channel into the ground truth heatmap which is jointly trained with other channels.

In summary, our \textbf{main contributions include:}
\begin{itemize}
%  \item Locate two issues with the MSE loss in heatmap regression.
  \vspace*{-0.05in}
  \item Propose a novel loss function for heatmap regression named Adaptive Wing loss, that is able to adapt its curvature to ground truth pixel values. This adaptive property reduces small errors on foreground pixels for accurate landmark localization, while tolerates small errors on background pixels for a better convergence rate. With proposed Weighted Loss Map it is also able to focus on foreground pixels and difficult background pixels during training.
  \vspace*{-0.05in}
  \item Encode coordinate information, including coordinates on boundary, into the face alignment algorithm using CoordConv~\cite{CoordConv}.
\end{itemize}
\vspace*{-0.1in}
Our approach outperforms the state-of-the-art algorithms by a significant margin on mainstream face alignment datasets including 300W~\cite{300W}, COFW~\cite{burgos2013robust} and WFLW~\cite{LAB}. We also show the validity of the Adaptive Wing loss in the human pose estimation task which also utilizes heatmap regression.
%-------------------------------------------------------------------------
\vspace*{-0.05in}
\section{Related Work}
\vspace*{-0.05in}
\textbf{CNN based heatmap regression models} leverage CNN to perform heatmap regression. In recent work~\cite{heatmap_regre_3, tang2018quantized, bulat2017binarized, bulat2017far}, joint bottom-up and top-down architectures such as stacked HG~\cite{Hourglass} were able to achieve the state-of-the-art performance. Bulat \etal~\cite{bulat2017binarized} proposed a hierarchical, parallel and multi-scale block as a replacement for the original ResNet~\cite{resnet} block to further improve the localization accuracy of HG. Tang \etal~\cite{tang2018quantized} was able to achieve current state-of-the-art with quantized densely connected U-Nets with fewer parameters than stacked HG models. Other architectures are also able to achieve excellent performance. Merget \etal~\cite{merget2018robust} proposed a fully convolutional neural network (FCN) that combines global and local context information for a refined prediction. Valle \etal~\cite{valle2018deeply} combined CNN with ensemble of regression trees in a coarse-to-fine fashion to achieve the state-of-the art accuracy. Another focus of this area is the 3D face alignment ~\cite{pose-invariant-face-alignment-with-a-single-cnn, dense-face-alignment}, that aims to provide 3D dense alignment based on 2D images.

\textbf{Loss functions for heatmap regression} were rarely studied in previous work. GoDP~\cite{GoDP} used a distance-aware softmax loss to assign large penalty on incorrectly classified positive samples, while gradually reducing penalty on miss-classified negative samples as the distance from nearby positive samples decrease. The Wing loss~\cite{Feng_Wing_Loss} is a modified log loss for direct regression of landmark coordinates. Compared with MSE, it amplifies the influence of small errors. Although the Wing loss is able to achieve the state-of-the-art performance in coordinate regression, it is not applicable to heatmap regression due to its high sensitivity to small errors on background pixels and the discontinuity of gradient at zero. Our proposed Adaptive Wing loss is novel since it is able to adapt its curvature to different ground truth pixel values, such that it can be sensitive to small errors on foreground pixels yet be able to tolerance small errors on background pixels. Hence, our loss can be applied to heatmap regression while the original Wing loss cannot be.

\textbf{Boundary information} was first introduced into face alignment by Wu~\etal~\cite{LAB}. LAB proposed a two-stage network with a stacked HG model to generate a facial boundary map, and then regress facial landmark coordinates directly with the help of boundary map. We believe including boundary information is beneficial to the heatmap regression and utilized a modified version to our model.

\textbf{Coordinate Encoding.} Translation invariance is intrinsic to the convolution operation. Although CNN greatly benefited from this parameter sharing scheme, Liu \etal~\cite{CoordConv} showed the inability of the convolution operation to handle simple coordinate transforms, and proposed a new operation called CoordConv, which encodes coordinate information as additional channels before convolution operation. CoordConv was shown to improve vision tasks such as object detection and generative modeling. For face alignment, the input images are always generated from a face detector with small variance on location and scale. These properties inspire us to include CoordConv to help CNN learn the relationship among facial landmarks based on their absolute locations. 
%------------------------------------------------------------------------
\vspace*{-0.04in}
\section{Our Model}
\vspace*{-0.04in}
Our model is based on the stacked HG architecture from Bulat \etal~\cite{bulat2017far} which improved over the original convolution block design from Newell \etal~\cite{Hourglass}.  For each HG, the output heatmap is supervised with the ground truth heatmap. We also added a sub-task of boundary prediction as an additional channel of the heatmap. Coordinate encoding is added before the first convolution layer of our network and before the first convolution block of each HG module. An overview of our model is shown in Figure~\ref{fig:model}. 
\vspace*{-0.1in}

\begin{figure}[]
\centering
\begin{subfigure}{0.53\linewidth}
\centering
\includegraphics[width=1.0\linewidth]{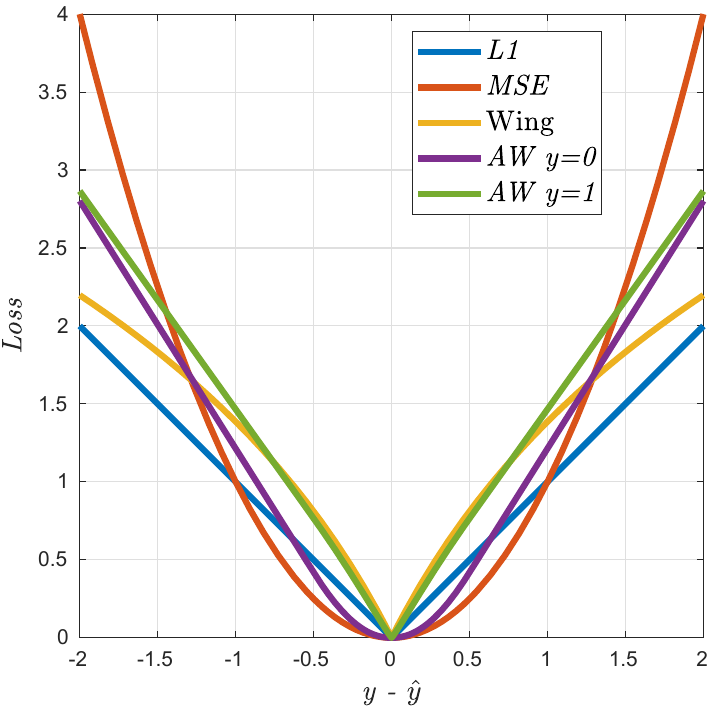}
\caption{Loss functions}
\label{fig:3D Wing sub1}
\end{subfigure}%
\begin{subfigure}{0.47\linewidth}
\includegraphics[width=1.0\linewidth]{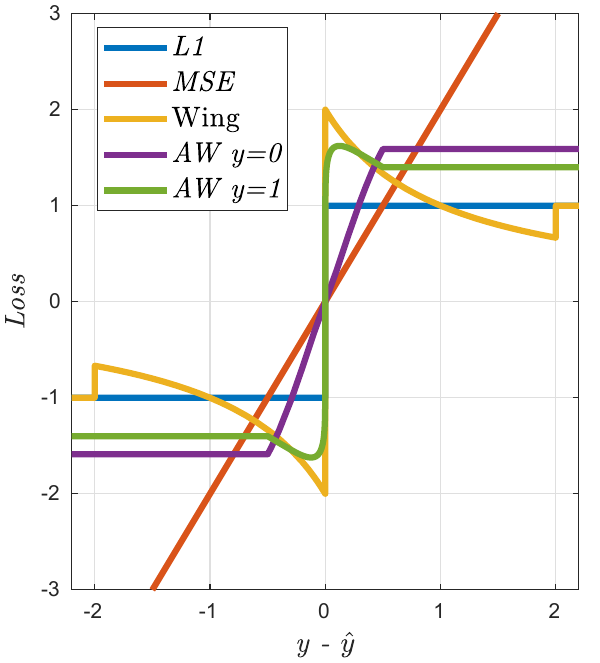}
\caption{Gradient of losses}
\label{fig:3D Wing sub2}
\end{subfigure}
\vspace*{-0.1in}
\caption{{\bf Different Loss Functions}. When $y=0$, the Adaptive Wing loss (purple) behaves similar to the MSE loss (red). When $y=1$, the Adaptive Wing loss (green) behaves similar to the Wing loss (yellow), but the gradient of the Adaptive Wing loss is smooth at point $y=\hat{y}$, as shown in Figure~\ref{fig:3D Wing sub2} (Best viewed in color).}
\label{fig: 3D Wing}
\vspace*{-0.2in}
\end{figure}

%------------------------------------------------------------------------
\section{Adaptive Wing Loss for Face Alignment}
%------------------------------------------------------------------------
\subsection{Loss function rationale}
\label{sec:loss_function_rationale}
Before starting our analysis, we would like to introduce a concept from robust statistics.  \textit{Influence}~\cite{hampel2011robust} is a heuristic tool used in robust statistics to investigate the properties of an estimator. In the context of our paper, the influence function is \textit{proportional to the gradient}~\cite{black1996unification} of our loss function. So if the gradient magnitude is large at point $y-\hat{y}$ (indicting the error), then we say the loss function has a large influence at point $y-\hat{y}$. If the gradient magnitude is close to zero at this point, then we say the loss function has a small influence at point $y-\hat{y}$. Theoretically, for heatmap regression, training is converged only if:
\vspace*{-0.05in}
\begin{equation}
\small
    \sum_{n=0}^{N}\sum_{i=0}^{H}\sum_{j=0}^{W}\sum_{k=0}^{C}\nabla Loss_{n}(y_{i,j,k}-\hat{y}_{i,j,k}) = 0
\end{equation}
where $N$ is the total number of training samples, $H$, $W$ and $C$ are the height, width and channels of heatmap, respectively. $Loss_n$ is the loss of $n-th$ sample, $y_{i,j,k}$ and $\hat{y}_{i,j,k}$ are ground truth pixel intensity and predicted pixel intensity respectively. At convergence, the influence of all errors must balance each other. Hence, a positive error on a pixel with large gradient magnitude (hence large influence) would need to be balanced by negative errors on many pixels with smaller influence. Errors with large gradient magnitude will also be more focused on during training compare to errors with small gradient magnitude.

The essence of heatmap regression is to output a Gaussian distribution centered at each ground truth landmark. Thus the accuracy of estimating pixel intensity at the mode of the Gaussian plays a vital role on correctly localizing landmarks. %During training stacked HG networks, we observed two issues with heatmap regression: i) Small errors do not have strong influence on loss function. This is the nature of the widely used MSE loss function. ii) During training all pixels are weighted equally in the loss function, so the error on background pixels will dominate in the backpropagation since they constitute more than 95\% of all the pixels. Hence the foreground pixels are not focused during training. 
The two issues we illustrated in Sec.~\ref{sec:introduction} result in an inaccurate estimation on the position of landmarks due to lacking of focus during training on foreground pixels. In this section and Sec.~\ref{sec:wing_loss}, we will discuss the causes of the first issue and how our proposed Adaptive Wing loss is able to remedy it. The second issue will be discussed in Sec.~\ref{sec:weighted_loss_map}.

The first issue is due to the commonly used MSE loss function for Heatmap regression. The gradient of the MSE loss is linear, so pixels with small errors have small influence, as shown in Figure~\ref{fig:3D Wing sub2}. This property could cause training to converge while many pixels still have small errors. As a result, models trained with MSE loss tend to predict a blurry and dilated heatmap. Even worse, the predicted heatmap often has low intensity on foreground pixels around difficult landmarks, e.g.\ occluded landmarks or faces with unusual illumination conditions. Accurately localizing landmarks from these low intensity pixels can be difficult. A good example can be found in Figure~\ref{fig:Heatmap_comparsion}.

L1 loss has constant gradient so that pixels with small errors have the same influence as pixels with large errors. However, the gradient of L1 loss is not continuous at point zero, which means for convergence, the amount of pixels with positive errors has to be exactly equal to the amount that has negative errors. The difficulty of achieving such delicate balance could cause training process to be unstable and oscillating.

Feng \etal~\cite{Feng_Wing_Loss} is able to improve the above loss functions by proposing Wing loss that has constant gradient when error is large, and large gradient when the error is small. Thus pixels with small errors will be amplified. The Wing loss is defined as follows:
\vspace*{-0.05in}
\begin{equation}
\small
Wing(y,\!\hat{y}\!)\! =\!
\begin{cases}
 \omega\! \ln(1\! +\!  \displaystyle |\frac{y\!-\!\hat{y}\!}{\epsilon}|)\! &\! \text{if }
|(y\!-\!\hat{y}\!)|\! <\! \omega   \\
  |y-\hat{y}| - C & \text{otherwise}
\end{cases}
\vspace*{-0.05in}
\end{equation}
where $y$ and $\hat{y}$ are the pixel values on ground truth heatmap and the predicted heatmap respectively, $C=\omega - \omega \ln(1 + \omega/\epsilon)$ is used to make function continuous at $|y - \hat{y}| = \omega$.
The Wing loss is, however, still not be able to overcome the discontinuity of its gradient at $y-\hat{y}=0$, with its large gradient magnitude around this point, training is even more difficult to converge compared with L1 loss. This property makes the Wing loss not applicable for heatmap regression, since with the Wing loss calculated on all background pixels, small errors on background pixels are having out-of-proportion influence. Training a neural network that outputs zero or small gradient on these pixels is very difficult. According to our experiment, the training of a heatmap regression network with the Wing loss is never able to converge. 

The above analysis leads us to define the desired properties of an ideal loss function for heatmap regression. We expect our loss function to have a constant influence when error is large, so that it will be robust to inaccurate annotations and occlusions. As the training process continues and errors getting smaller, there will be two scenarios: i) \textbf{For foreground pixels}, the influence (as well as the gradient) should start to increase so that the training is able to focus on reducing these errors. The influence should then decrease rapidly as the errors go very close to zero, so that these "good enough" pixels will no longer be focused on. The reduced influence of correct estimations helps the network to stay converged, instead of oscillating like the L1 and the Wing loss. ii) \textbf{For background pixels}, the gradient should behaves more similar to the MSE loss, that is, it will gradually decrease to zero as the training error decreases, so that the influence will be relatively small when the errors are small. This property reduces the focus of the training on background pixels, stabilizing the training process.

A fixed loss function cannot achieve both properties simultaneously. Thus, the loss function should be able to adapt to different pixel intensities on the ground truth heatmaps. As the ground truth pixels close to the mode (have intensities that are close to 1), the influence of small  errors should increase. With ground truth pixel intensities close to 0, the loss function should behave more similar to the MSE loss. Since pixel values on the ground truth heatmap range from 0 to 1, we also expect our loss function to have a smooth transition according to different pixel values.

%------------------------------------------------------------------------
\subsection{The Adaptive Wing Loss}
\label{sec:wing_loss}
\begin{figure}[]
\centering
\begin{subfigure}{0.5\linewidth}
\centering
\includegraphics[width=1.0\linewidth]{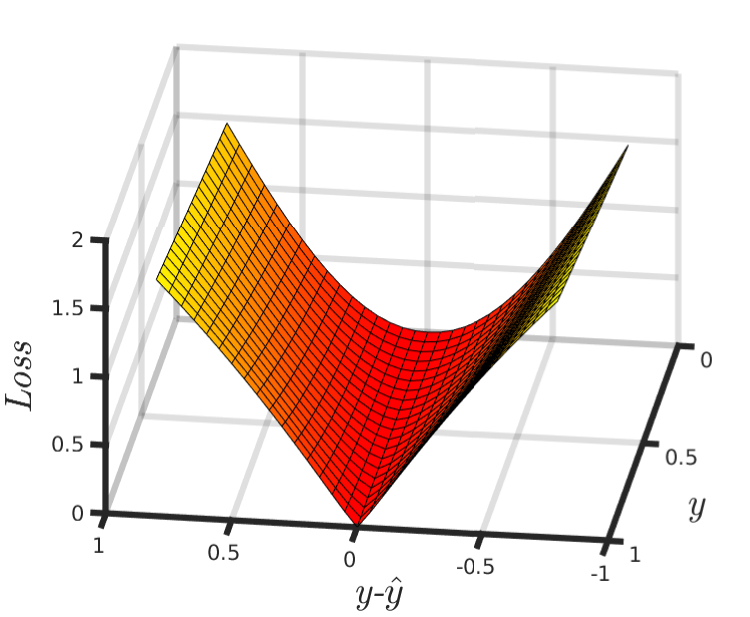}
\caption{AWing loss}
\label{fig:3d_awing_1}
\end{subfigure}%
\begin{subfigure}{0.5\linewidth}
\centering
\includegraphics[width=0.9\linewidth]{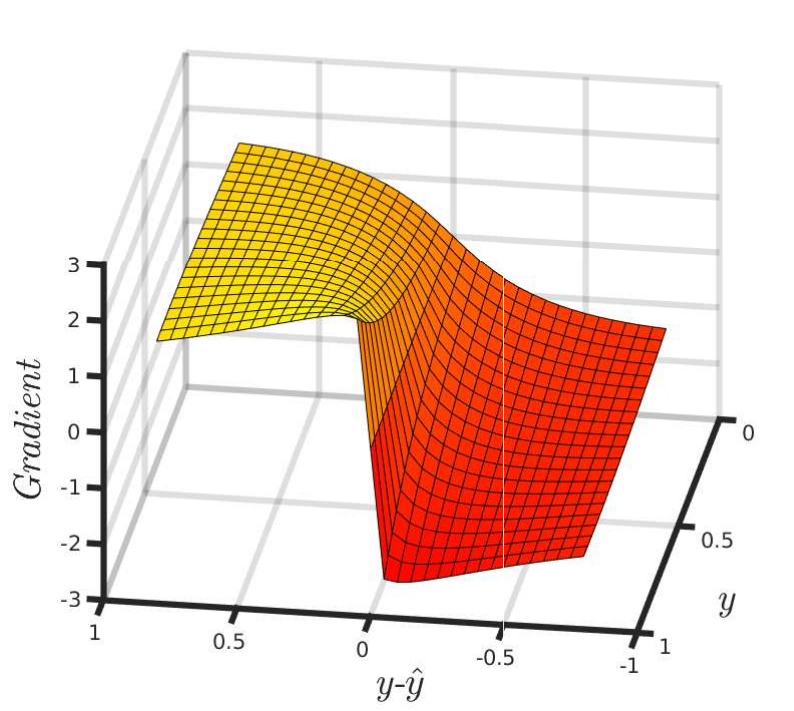}
\caption{Gradient of AWing}
\label{fig:3d_awing_2}
\end{subfigure}%
\vspace*{-0.1in}
\caption{{\bf The nonlinear part of the Adaptive Wing loss} is able to adapt its shape according to different values of $y$. As $y$ increases, the shape is more similar to the Wing loss, and the influence of small errors (near-side of the y axis) will remain strong. As $y$ decreases, the influence on these errors will decrease and the loss function will behave more like MSE.}
\label{fig:3d_awing}
\vspace*{-0.2in}
\end{figure}

Following intuitions above, we propose our Adaptive Wing (AWing) loss, defined as follows:
\vspace*{-0.05in}
\begin{equation}
\small
AWing(y,\!\hat{y})\! =\!
\begin{cases}
 \omega\! \ln(1\! +\!  \displaystyle |\frac{y\!-\!\hat{y}\!}{\epsilon}|^{\alpha-y})\! &\! \text{if }
|(y\!-\!\hat{y})|\! <\! \theta   \\
  A|y-\hat{y}\!| - C & \text{otherwise}
\end{cases}
\vspace*{-0.05in}
\end{equation}
where $y$ and $\hat{y}$ are the pixel values on the ground truth heatmap and the predicted heatmap respectively, $\omega, \theta, \epsilon$ and $\alpha$ are positive values, $A = \omega (1/(1+(\theta/\epsilon)^{(\alpha-y)}))(\alpha-y)((\theta/\epsilon)^{(\alpha-y-1)})(1/\epsilon)$ and $C = (\theta A - \omega\ln(1 + ( \theta/\epsilon )^{\alpha-y}))$ are used to make loss function continuous and smooth at $|y - \hat{y}| = \theta$. Unlike Wing loss which uses $\omega$ as the threshold, we introduce a new variable $\theta$ as a threshold to switch between linear and nonlinear part. For heatmap regression, we often regress a value between 0 and 1, so we expect our threshold lies in this range. When $|y-\hat{y}|<\theta$, we consider the error to be small and need stronger influence. More importantly, we adopt an exponential term $\alpha-y$, which is used to adapt the shape of the loss function to $y$ and makes loss function smooth at point zero. Note $\alpha$ has to be slightly larger than 2 to maintain the ideal properties we discussed in Sec.~\ref{sec:loss_function_rationale}, this is due to the normalization of $y$ in the range of $[0, 1]$. For pixels on $y$ with values close to 1 (the landmarks we want to localize), the power term $\alpha -y$ will be slightly larger than 1, and the nonlinear part will behave like Wing loss, which has large influence on smaller errors. But different from Wing loss, the influence will decrease to zero rapidly as errors are very close to zero (see Fig.~\ref{fig: 3D Wing}). As $y$ decreases, the loss function will shift to a MSE-like loss function, which allows the training not to focus on the pixels that still have errors but small influence. Figure~\ref{fig:3d_awing} shows how the power term $\alpha - y$ facilities the smooth transition across different values of $y$, so that the influence of small errors will gradually increase as the value of $y$ increases. Larger $\omega$ and smaller $\epsilon$ values will increase the influence on small errors and vice versa, large $\omega$ values are shown to be effective according to our experiment.

The nonlinear part of our Adaptive Wing loss function behaves similarly to Lorentzian (aka. Cauchy) loss~\cite{black1996robust} in a more generalized fashion. But different from robust loss functions such as Lorentzian and Geman-McClure~\cite{ganan1985bayesian}, we do not need the gradient to decrease to zero as error increases. This is due to the nature of heatmap regression. In robust regression, the learner learns to ignore noisy outliers with large error. In the context of face alignment, all facial landmarks are annotated with relatively small noises, so we do not have noisy outliers to ignore. A linear loss is sufficient for the training to converge to a location where predictions will be fairly close to the ground truth heatmap, and after that the loss function will switch to its nonlinear part to refine the prediction with increased influence on small errors. In practice, we found the linear form when errors are large to achieve better performance, compared with keep using the nonlinear form when the error is large.

We empirically used $\alpha=2.1$ in our model. In our experiments, we found $\omega=14$, $\epsilon=1$, $\theta=0.5$ to be most effective, detailed ablation studies on parameter settings are shown at Sec.~\ref{sec:aw_parameters}.

%------------------------------------------------------------------------
\subsection{Weighted loss map}
\label{sec:weighted_loss_map}
In this section we will discuss the second issue in Sec.~\ref{sec:loss_function_rationale}. In a typical setting for facial landmark localization with a $64\times64$ heatmap, and the size of Gaussian of $7\times7$, foreground pixels only constitute 1.2\% of all the pixels. Assigning equal weight for such an unbalanced data could make the training process slow to converge and result in an inferior performance. To further establish the network's ability to focus on foreground pixels and difficult background pixels (background pixels that are close to foreground pixels), we introduce the Weighted Loss Map to balance the loss from different types of pixels. We first define our loss map mask to be:
\vspace*{-0.05in}
\begin{equation}
    \small
    M = 
    \begin{cases}
        1 & \text{where } H^{d} > = 0.2 \\
        0 & \text{otherwise}
    \end{cases}
    \vspace*{-0.05in}
\end{equation}
where $H^{d}$ is generated from ground truth heatmap $H$ by a $3\times3$ gray dilation. The loss map mask $M$ assigns foreground pixels and difficult background pixels 1, and other pixels 0.

\begin{figure}[H]
\centering
\begin{subfigure}{0.3\linewidth}
\centering
\includegraphics[width=0.9\linewidth]{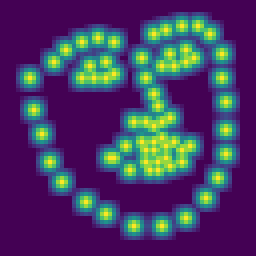}
\caption{$H$}
\label{fig:GT_gray}
\end{subfigure}%
\begin{subfigure}{0.3\linewidth}
\centering
\includegraphics[width=0.9\linewidth]{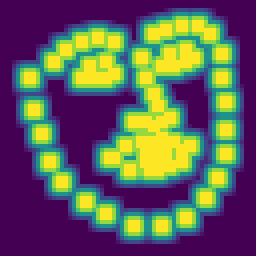}
\caption{$H^{d}$}
\label{fig:GT_dilated}
\end{subfigure}%
\begin{subfigure}{0.3\linewidth}
\centering
\includegraphics[width=0.9\linewidth]{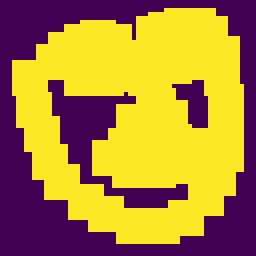}
\caption{$M$}
\label{fig:GT_binarized}
\end{subfigure}%
\vspace*{-0.1in}
\caption{Important pixels are generated by dilating $H$ from Figure~\ref{fig:GT_gray} with 3x3 dilation, and then binarizing to Figure~\ref{fig:GT_binarized} with a threshold of 0.2. For visualization purposes, all channels are max-pooled into one channel.}
\label{fig: heatmap_dilation}
\vspace*{-0.1in}
\end{figure}

With the loss map mask $M$, We define our Weighted Loss Map as follows:
\vspace*{-0.05in}
\begin{equation}
    \small
    Loss_{weighted}(H, \hat{H}) = Loss(H, \hat{H})\otimes (W\cdot M+1)
    \vspace*{-0.05in}
\end{equation}
where $\otimes$ is element-wise production, $W$ is a scalar hyperparameter to control how much weight to be added. See Figure~\ref{fig: heatmap_dilation} for a visualization of weight map generation. In our experiments we use $W=10$.
The intuition is to assign pixels on heatmap with different weights. Foreground pixels have to be focused on during training, since these pixels are the most useful for localizing the mode of the Gaussian distribution. Difficult background pixels should also be focused on since these pixels are relatively difficult to regress, accurately regressing them could help narrow down the area of foreground pixels to improve localization accuracy.

%-------------------------------------------------------------------------
\section{Boundary Information}
Inspired by \cite{LAB}, we introduce boundary prediction into our network as a sub-task, but in a different manner. Instead of breaking boundaries into different parts, we use only one additional channel as the boundary channel that combines all boundary lines to our heatmap. We believe this will efficiently capture the global information on a human face. The boundary information then will be aggregated into the network naturally via convolution operations in a forward pass, and will also be used in Section \ref{sec:coord aggregation} to generate the boundary coordinate map, which can further improve localization accuracy according to our ablation study in Sec.~\ref{sec:ablation_study}.

%-------------------------------------------------------------------------
\section{Coordinate aggregation}
\label{sec:coord aggregation}
We integrate CoordConv~\cite{CoordConv} into our model to improve the capability of traditional convolutional neural network to capture coordinate information. In addition to $X$, $Y$ and radius coordinate encoding in ~\cite{CoordConv}, we also leverage our boundary prediction to generate $X$ and $Y$ coordinates only at boundary. More specifically, we define $X$ coordinate encoding to be $C_{x}$, the boundary prediction from previous HG is $B$, the boundary coordinate encoding $B_{x}$ is defined as:
\vspace*{-0.05in}
\begin{equation}
\small
    B_{x} =
    \begin{cases}
        C_{x} & \text{where } B > = 0.05 \\
        0 & \text{otherwise}
    \end{cases}
    \vspace*{-0.05in}
\end{equation}
$B_{y}$ is generated in the similar fashion from $C_{y}$. The  coordinate channels are generated at runtime and then concatenated with the original input to perform regular convolution.

%-------------------------------------------------------------------------
\section{Experiments}

%-------------------------------------------------------------------------
\subsection{Datasets}
%-------------------------------------------------------------------------
We tested our approach on the \textbf{COFW}~\cite{burgos2013robust}, \textbf{300W~\cite{300W}}, 300W private test dataset and the \textbf{WFLW~\cite{LAB}} dataset. The WFLW dataset is the most difficult dataset of them all. For more details on theses datasets, please refer to supplementary materials.

\subsection{Evaluation Metrics}
\textbf{Normalized Mean Error (NME)} is commonly used to evaluate the quality of face alignment algorithms. The NME for each image is defined as:
\vspace*{-0.05in}
\begin{small}
\begin{equation}
    NME(P, \hat{P}) = \frac{1}{M}\sum_{i=1}^{M}\frac{||p_{i} - \hat{p_{i}}||_2}{d}
    \vspace*{-0.05in}
\end{equation}
\end{small}
where $P$ and $\hat{P}$ are the ground truth and the predicted landmark coordinates for each image respectively, $M$ is the number of landmarks of each image, $\hat{p_i}$ is the i-th predicted landmark coordinates in $P$ and $p_i$ is the i-th ground truth landmark coordinates in $\hat{P}$, $d$ is the normalization factor. For the COFW dataset, we use inter-pupil (distance of eye centers) as the normalization factor. For the 300W dataset, we provide both inter-ocular distance (distance of outer eye corners) used as the original evaluation protocol in ~\cite{300W}, and inter-pupil distance used in \cite{300W_inter_pupil}. For the WFLW dataset, we use the inter-ocular distance described in ~\cite{LAB}.

\textbf{Failure Rate (FR)} is another metric to evaluate localization quality. For one image, if NME is larger than a threshold, then it is considered a failed prediction. For the 300W private test dataset, we use $8\%$ and $10\%$ respectively to compare with different approaches. For the WFLW dataset, we follow~\cite{Feng_Wing_Loss, LAB} and use $10\%$ as the threshold. 

\textbf{Cumulative Error Distribution (CED)} curve shows the NME to the proportion of total test samples. The curve is usually plotted from zero up to the NME failure rate threshold (e.g.\ $10\%$, $8\%$). Area Under Curve (AUC) is calculated based on the CED curve. Larger AUC reflects that larger portion of the test dataset is well predicted.

%-------------------------------------------------------------------------
\subsection{Implementation details}
During training and testing, we use provided bounding boxes from dataset (with the longer side as the length of a square) to crop faces from images, except for the 300W private test dataset since no official bounding boxes are provided. For the WFLW dataset, the provided bounding boxes are not very accurate, to ensure all landmarks are preserved from cropping, we enlarge the bounding boxes by 10\% on both dimensions. For the 300W private test dataset, we use ground truth landmarks to crop faces.

The input of the network is $256\times 256$, the output of each stacked HG is $64\times 64$. We use four stacks of HG, same with other baselines. During training, we use RMSProp~\cite{RMSProp} with an initial learning rate of $1\times 10^{-4}$. We set the momentum to be 0 (adopted from~\cite{bulat2017far, Hourglass}) and the weight decay to be $1\times 10^{-5}$. We train for 240 epoches, and the learning rate is reduced to $1\times 10^{-5}$ and $1\times 10^{-6}$ after 80 and 160 epoches. Data augmentation is performed with random rotation ($\pm50\degree$), translation ($\pm25px$), flipping ($50\%$), and rescaling ($\pm15\%$). Random Gaussian blur, noise and occlusion are also used. All models are trained from scratch. During inference, we adopt the same strategy used in Newell \etal~\cite{Hourglass}, the location on the pixel with the highest response is shifted a quarter pixel to the second highest nearby pixel. The boundary line is generated from landmarks via distance transform similar to ~\cite{LAB}, different boundary lines are merged into one channel by selecting maximum values on each pixel across all channels.

\begin{table*}
\vspace*{-0.1in}
\centering
\resizebox{0.7\linewidth}{!}{%
\begin{tabular}{|ccccccccc|}
\hline
Metric                            & Method & Testset & \begin{tabular}[c]{@{}c@{}}Pose\\ Subset\end{tabular} & \begin{tabular}[c]{@{}c@{}}Expression\\ Subset\end{tabular} & \begin{tabular}[c]{@{}c@{}}Illumination\\ Subset\end{tabular} & \begin{tabular}[c]{@{}c@{}}Make-up\\ Subset\end{tabular} & \begin{tabular}[c]{@{}c@{}}Occlusion\\ Subset\end{tabular} & \begin{tabular}[c]{@{}c@{}}Blur\\ Subset\end{tabular} \\ \hline
\multirow{7}{*}{NME(\%)}   & ESR\textsubscript{CVPR 14}~\cite{cao2014face}   & 11.13  & 25.88  & 11.47  & 10.49  & 11.05  & 13.75  & 12.20  \\ 
                                  & SDM\textsubscript{CVPR 13}~\cite{xiong2013supervised}   & 10.29  & 24.10  & 11.45  & 9.32   & 9.38   & 13.03  & 11.28  \\
                                  & CFSS\textsubscript{CVPR 15}~\cite{zhu2015face}  & 9.07   & 21.36  & 10.09  & 8.30   & 8.74   & 11.76  & 9.96   \\
                                  & DVLN\textsubscript{CVPR 17}~\cite{wu2017leveraging}  & 6.08   & 11.54  & 6.78   & 5.73   & 5.98   & 7.33   & 6.88   \\
                                  & LAB\textsubscript{CVPR 18}~\cite{LAB}   & 5.27   & 10.24  & 5.51   & 5.23   & 5.15   & 6.79   & 6.32   \\
                                  & Wing\textsubscript{CVPR 18}~\cite{Feng_Wing_Loss}  & 5.11   & 8.75   & 5.36   & 4.93   & 5.41   & 6.37   & 5.81   \\
                                  & \textbf{AWing(Ours)} & \textbf{4.36} & \textbf{7.38} & \textbf{4.58} & \textbf{4.32} & \textbf{4.27} & \textbf{5.19} & \textbf{4.96} \\ 
                                  & \textbf{AWing(GTBbox)} & 4.21 & 7.21 & 4.46 & 4.23 & 4.02 & 4.99 & 4.82 \\ \hline
\multirow{7}{*}{FR\textsubscript{10\%}(\%)} & ESR\textsubscript{CVPR 14}~\cite{cao2014face}   & 35.24  & 90.18  & 42.04  & 30.80  & 38.84  & 47.28  & 41.40  \\
                                  & SDM\textsubscript{CVPR 13}~\cite{xiong2013supervised}   & 29.40  & 84.36  & 33.44  & 26.22  & 27.67  & 41.85  & 35.32  \\
                                  & CFSS\textsubscript{CVPR 15}~\cite{zhu2015face}  & 20.56  & 66.26  & 23.25  & 17.34  & 21.84  & 32.88  & 23.67  \\
                                  & DVLN\textsubscript{CVPR 17}~\cite{wu2017leveraging}  & 10.84  & 46.93  & 11.15  & 7.31   & 11.65  & 16.30  & 13.71  \\
                                  & LAB\textsubscript{CVPR 18}~\cite{LAB}   & 7.56   & 28.83  & 6.37   & 6.73   & 7.77   & 13.72  & 10.74  \\
                                  & Wing\textsubscript{CVPR 18}~\cite{Feng_Wing_Loss}  & 6.00   & 22.70  & 4.78   & 4.30   & 7.77   & 12.50  & 7.76   \\
                                  & \textbf{AWing(Ours)} & \textbf{2.84} & \textbf{13.50} & \textbf{2.23} & \textbf{2.58} & \textbf{2.91} & \textbf{5.98} & \textbf{3.75}   \\
                                  & \textbf{AWing(GTBbox)} & 2.04 & 9.20 & 1.27 & 2.01 & 0.97 & 4.21 & 2.72   \\ \hline
\multirow{7}{*}{AUC\textsubscript{10\%}}              & ESR\textsubscript{CVPR 14}~\cite{cao2014face}   & 0.2774 & 0.0177 & 0.1981 & 0.2953 & 0.2485 & 0.1946 & 0.2204 \\
                                  & SDM\textsubscript{CVPR 13}~\cite{xiong2013supervised}   & 0.3002 & 0.0226 & 0.2293 & 0.3237 & 0.3125 & 0.2060 & 0.2398 \\
                                  & CFSS\textsubscript{CVPR 15}~\cite{zhu2015face}  & 0.3659 & 0.0632 & 0.3157 & 0.3854 & 0.3691 & 0.2688 & 0.3037 \\
                                  & DVLN\textsubscript{CVPR 17}~\cite{wu2017leveraging}  & 0.4551 & 0.1474 & 0.3889 & 0.4743 & 0.4494 & 0.3794 & 0.3973 \\
                                  & LAB\textsubscript{CVPR 18}~\cite{LAB}   & 0.5323 & 0.2345 & 0.4951 & 0.5433 & 0.5394 & 0.4490 & 0.4630 \\
                                  & Wing\textsubscript{CVPR 18}~\cite{Feng_Wing_Loss}  & 0.5504 & 0.3100 & 0.4959 & 0.5408 & 0.5582 & 0.4885 & 0.4918 \\
                                  & \textbf{AWing(Ours)} & \textbf{0.5719} & \textbf{0.3120}   & \textbf{0.5149} & \textbf{0.5777} & \textbf{0.5715} & \textbf{0.5022} & \textbf{0.5120} \\ 
                                  & \textbf{AWing(GTBbox)} & 0.5895 & 0.3337   & 0.5718 & 0.5958 & 0.6017 & 0.5275 & 0.5393 \\ \hline
\end{tabular}}
\vspace*{-0.1in}
\caption{Evaluation on the WFLW dataset. GTBbox indicates the ground truth landmarks are used to crop faces.}
\label{table:WFLW}
\end{table*}

\begin{table}[H]
\vspace*{-0.05in}
\centering
\setlength\tabcolsep{3pt}
\resizebox{0.7\linewidth}{!}{%
\begin{tabular}{|cccc|}
\hline
Method              & NME       & AUC\textsubscript{10\%}       & FR\textsubscript{10\%}        \\ \hline
Human~\cite{burgos2013robust}                 & 5.60         & -     & 0.00     \\
TCDCN\textsubscript{ECCV 14}~\cite{zhang2014facial}               & 8.05         & -     & -     \\
Wu \etal\textsubscript{ICCV 15}~\cite{wu2015robust}        & 5.93      & -     & -      \\
RAR\textsubscript{ECCV 16}~\cite{xiao2016robust}                & 6.03         & -     & 4.14     \\
DAC-CSR\textsubscript{CVPR 17}~\cite{feng2017dynamic}       & 6.03      & -     & 4.73      \\
SHN\textsubscript{CVPRW 17}~\cite{yang2017stacked}         & 5.60      & -         & -         \\ 
PCD-CNN\textsubscript{CVPR 18}~\cite{kumar2018disentangling}   & 5.77      & -         & 3.73         \\
Wing\textsubscript{CVPR 18}~\cite{Feng_Wing_Loss}  & 5.44 & - &	3.75 \\ \hline
\textbf{AWing(Ours)}      & \textbf{4.94} & \textbf{48.82} & \textbf{0.99} \\ \hline
                    & NME       & AUC\textsubscript{8\%}       & FR\textsubscript{8\%}        \\ \hline
DCFE\textsubscript{ECCV 18}~\cite{valle2018deeply}     & 5.27          & 35.86    & 7.29      \\ \hline
\textbf{AWing(Ours)}      & \textbf{4.94} & \textbf{39.11} & \textbf{5.52} \\ \hline
\end{tabular}}
\vspace*{-0.1in}
\caption{Evaluation on the COFW dataset}
\vspace*{-0.2in}
\label{table:cofw}
\end{table}

\subsubsection{Evaluation on COFW}
Experiment results on the COFW dataset is shown in Table~\ref{table:cofw}. Our approach outperforms previous state-of-the-art by a significant margin, especially on the failure rate. We are able to reduce the failure rate measured at 10\% NME from 3.73\% to 0.99\%. As for NME, our method perform much better than human (5.60\%). Our performance on the COFW shows the robustness of our approach against faces with large pose and heavy occlusion.

%-------------------------------------------------------------------------
\subsection{Evaluation on 300W}
Our method is able to achieve the state-of-the-art performance on the 300W testing dataset, see Table~\ref{table:300W}. For the challenge subset (iBug dataset), we are able to outperform Wing~\cite{Feng_Wing_Loss} by a significant margin, which also proves the robustness of our approach against occlusion and large pose variation.  %, %but underperform it (with a much simpler network CNN6/7 as backbone) on Common Subset and Fullset , which we argue the performance reported on 300W dataset in Feng~\etal~\cite{Feng_Wing_Loss} needs to be rechecked.
Furthermore, on the 300W private test dataset (Table~\ref{table:300W_challenge}), we again outperform the previous state-of-the-art on variant metrics including NME, AUC and FR measured with either 8\% NME and 10\% NME. Note that we more than halved the failure rate of the next best baseline to 0.83\%, which means only 5 faces out of 600 have an NME that is larger than 8\%.

\begin{table}[]
\vspace*{-0.15in}
\centering
\setlength\tabcolsep{2.5pt}
\resizebox{0.85\linewidth}{!}{%
\begin{tabular}{|cccc|}
\hline
Method  & \begin{tabular}[c]{@{}c@{}}Common\\ Subset\end{tabular} & \begin{tabular}[c]{@{}c@{}}Challenging\\ Subset\end{tabular} & Fullset \\ \hline
\multicolumn{4}{|c|}{Inter-pupil Normalization}                                                                                             \\ \hline 
%RCPR\textsubscript{CVPR 13} \cite{burgos2013robust}    & 6.18    & 17.26     & 8.35    \\ 
CFAN\textsubscript{ECCV 14}~\cite{zhang2014coarse}    & 5.50    & 16.78     & 7.69    \\
%ESR\textsubscript{CVPR 14}~\cite{cao2014face}     & 5.28    & 17.00     & 7.58    \\
SDM\textsubscript{CVPR 13}~\cite{xiong2013supervised}     & 5.57    & 15.40     & 7.50    \\
LBF\textsubscript{CVPR 14}~\cite{ren2014face}     & 4.95    & 11.98     & 6.32    \\
CFSS\textsubscript{CVPR 15}~\cite{zhu2015face}    & 4.73    & 9.98      & 5.76    \\
%3DDFA\textsubscript{CVPR 16}~\cite{zhu2016face}   & 6.15    & 10.59     & 7.01    \\
TCDCN\textsubscript{16'}~\cite{zhang2016learning}   & 4.80    & 8.60      & 5.54    \\
MDM\textsubscript{CVPR 16}~\cite{trigeorgis2016mnemonic}     & 4.83    & 10.14     & 5.88    \\
RAR\textsubscript{ECCV 16}~\cite{xiao2016robust}     & 4.12    & 8.35      & 4.94    \\
DVLN\textsubscript{CVPR 17}~\cite{wu2017leveraging}    & 3.94    & 7.62      & 4.66    \\
TSR\textsubscript{CVPR 17}~\cite{cnn_direct_3}           & 4.36      & 7.56       & 4.99    \\
DSRN\textsubscript{CVPR 18}~\cite{miao2018direct}        & 4.12      & 9.68       & 5.21    \\
\small{RCN\textsuperscript{+}(L+ELT)\textsubscript{CVPR 18})~\cite{honari2018improving}} & 4.20 & 7.78  & 4.90  \\
DCFE\textsubscript{ECCV 18}~\cite{valle2018deeply}      & 3.83        & 7.54      & 4.55 \\
LAB\textsubscript{CVPR 18}~\cite{LAB}                    & 3.42       &  6.98       & 4.12    \\
Wing\textsubscript{CVPR 18}~\cite{Feng_Wing_Loss}  & \textbf{3.27}  & 7.18      & \textbf{4.04} \\ \hline
\textbf{AWing(Ours)}                                    & 3.77     & \textbf{6.52}   & 4.31     \\ \hline
\multicolumn{4}{|c|}{Inter-ocular Normalization}                                                       \\ \hline
PCD-CNN\textsubscript{CVPR 18}~\cite{kumardisentangling} & 3.67    & 7.62        & 4.44    \\
CPM+SBR\textsubscript{CVPR 18}~\cite{dong2018style}  &3.28         & 7.58        & 4.10    \\
SAN\textsubscript{CVPR 18}~\cite{dong2018style}     & 3.34         & 6.60         & 3.98    \\
LAB\textsubscript{CVPR 18}~\cite{LAB}                & 2.98         & 5.19        & 3.49    \\
DU-Net\textsubscript{ECCV 18}~\cite{tang2018quantized}  & 2.90      & 5.15        & 3.35    \\ \hline
\textbf{AWing(Ours)}                               & \textbf{2.72}  & \textbf{4.52}   & \textbf{3.07}        \\ \hline
\end{tabular}}
\caption{Evaluation on the 300W testset}
\label{table:300W}
\vspace*{-0.15in}
\end{table}

\begin{table}[]
\vspace*{-0.15in}
\centering
\setlength\tabcolsep{3pt}
\resizebox{0.7\linewidth}{!}{%
\begin{tabular}{|cccc|}
\hline
Method              & NME       & AUC\textsubscript{8\%}       & FR\textsubscript{8\%}        \\ \hline
ESR\textsubscript{CVPR 14}~\cite{cao2014face}                 & -         & 32.35     & 17.00     \\
cGPRT\textsubscript{CVPR 15}~\cite{lee2015face}               & -         & 41.32     & 12.83     \\
CFSS\textsubscript{CVPR 15}~\cite{zhu2015face}                & -         & 39.81     & 12.30     \\
MDM\textsubscript{CVPR 16}~\cite{trigeorgis2016mnemonic}                 & 5.05      & 45.32     & 6.80      \\
DAN\textsubscript{CVPRW 17}~\cite{heatmap_regre_2}                 & 4.30      & 47.00     & 2.67      \\
SHN\textsubscript{CVPRW 17}~\cite{heatmap_regre_3}                 & 4.05      & -         & -         \\
DCFE\textsubscript{ECCV 18}~\cite{valle2018deeply}                & 3.88      & 52.42     & 1.83      \\ \hline
\textbf{AWing(Ours)}      & \textbf{3.56} & \textbf{55.76} & \textbf{0.83} \\ \hline
                    & NME       & AUC\textsubscript{10\%}       & FR\textsubscript{10\%}        \\ \hline
M3-CSR\textsubscript{16'}~\cite{deng2016m3}         & -          & 47.52    & 5.5       \\
Fan \etal \textsubscript{16'}~\cite{fan2016approaching} & -          & 48.02    & 14.83     \\
DR + MDM \textsubscript{CVPR 17}~\cite{guler2017densereg}      & -          & 52.19    & 3.67      \\
JMFA\textsubscript{17'}~\cite{deng2017joint}                & -          & 54.85    & 1.00      \\
LAB\textsubscript{CVPR 18}~\cite{LAB}                 & -          & 58.85    & 0.83      \\ \hline
\textbf{AWing(Ours)}      & \textbf{3.56} & \textbf{64.40} & \textbf{0.33} \\ \hline
\end{tabular}}
\vspace*{-0.1in}
\caption{Evaluation on the 300W private dataset}
\vspace*{-0.1in}
\label{table:300W_challenge}
\end{table}

%-------------------------------------------------------------------------
\subsection{Evaluation on WFLW}

\begin{figure}[]
\centering
\captionsetup[subfigure]{justification=centering}
\begin{subfigure}{0.2\linewidth}
\centering
\includegraphics[width=1.0\linewidth]{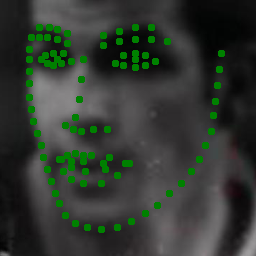}
\end{subfigure}%
\begin{subfigure}{0.2\linewidth}
\centering
\includegraphics[width=1.0\linewidth]{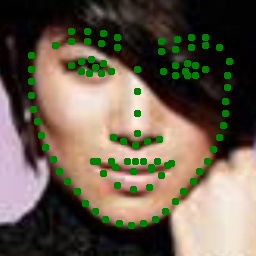}
\end{subfigure}%
\begin{subfigure}{0.2\linewidth}
\centering
\includegraphics[width=1.0\linewidth]{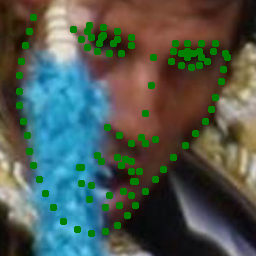}
\end{subfigure}%
\begin{subfigure}{0.2\linewidth}
\centering
\includegraphics[width=1.0\linewidth]{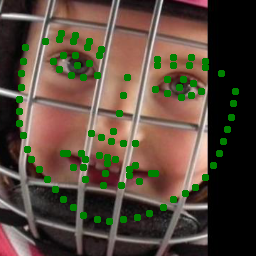}
\end{subfigure}%

\begin{subfigure}{0.2\linewidth}
\centering
\includegraphics[width=1.0\linewidth]{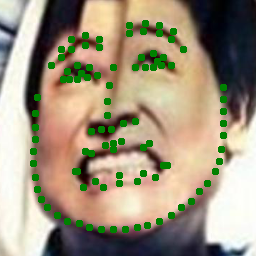}
\end{subfigure}%
\begin{subfigure}{0.2\linewidth}
\centering
\includegraphics[width=1.0\linewidth]{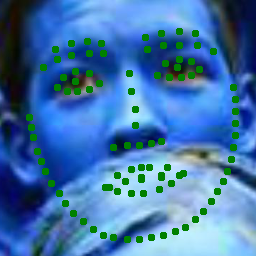}
\end{subfigure}%
\begin{subfigure}{0.2\linewidth}
\centering
\includegraphics[width=1.0\linewidth]{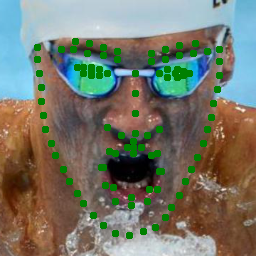}
\end{subfigure}%
\begin{subfigure}{0.2\linewidth}
\centering
\includegraphics[width=1.0\linewidth]{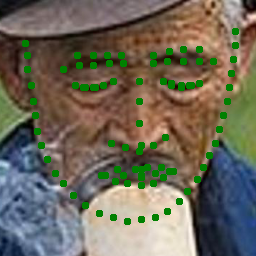}
\end{subfigure}%
\vspace*{-0.1in}
\caption{Visualizations on WFLW test dataset.}
\label{fig:wflw_vis}
\vspace*{-0.1in}
\end{figure}

Our method again achieves the best results on the WFLW dataset in Table~\ref{table:WFLW}, which is significantly more difficult than COFW and 300W (see Fig.~\ref{fig:wflw_vis} for visualizations). On every subset we outperform the previous state-of-the-art approaches by a significant margin. Note that the baseline Wing is using ResNet50~\cite{resnet} as the backbone architecture, which already performs better than the CNN6/7 architecture they used in COFW and 300W.
We are also able to reduce the failure rate and increase the AUC dramatically and hence improving the overall localization quality significantly. All in all, our approach fails on only 2.84\% of all images, more than a two times improvement compared with previous best results.

\vspace*{-0.1in}
%-------------------------------------------------------------------------
\subsection{Ablation study}
%-------------------------------------------------------------------------
\subsubsection{Evaluation on different loss function parameters}
\label{sec:aw_parameters}
To find the optimal parameter settings for the Adaptive Wing loss for heatmap regression, we examined different parameter combinations and evaluated on the WFLW dataset with faces cropped from ground truth landmarks. However, the search space is too large and we only have limited resources. To reduce the search space, we set our initial $\theta$ to 0.5, since the pixel value of the ground truth heatmap is from 0 to 1, we believe focusing on errors that are smaller than 0.5 is more than enough. Table~\ref{table:3D Wing Exps} shows NMEs on different combinations of $\omega$ and $\epsilon$. As a result, we picked $\omega=14$ and $\epsilon=1$. The experiments also show our Adaptive Wing loss is not very sensitive to $\omega$ and $\epsilon$, since the difference of NMEs are not significant within a certain range of different settings. Then we fixed $\omega$ and $\epsilon$, and examine different $\theta$, the results are shown in Table~\ref{table:different_theta}.
\label{sec:ablation_study}
\begin{table}[H]
\vspace*{-0.1in}
\centering
\resizebox{0.7\linewidth}{!}{%
\begin{tabular}{|c|ccccc|}
\hline
\backslashbox{$\epsilon$}{$\omega$} & 10 & 12 & 14 & 16 & 18 \\ \hline
0.5                  & 4.28  & 4.25  & 4.24  & 4.28  & 4.29   \\ \hline
1                & 4.24  & 4.26  & \textbf{4.21}  & 4.22  & 4.26    \\ \hline
2                  & 4.23  & 4.27   & 4.26  & 4.28  & 4.30   \\ \hline
\end{tabular}}
\vspace*{-0.1in}
\caption{Evaluation on different parameter settings of the Adaptive Wing loss.}
\label{table:3D Wing Exps}
\vspace*{-0.1in}
\end{table}

\begin{table}[H]
\vspace*{-0.1in}
\centering
\resizebox{0.7\linewidth}{!}{%
\begin{tabular}{|c|ccccc|}
\hline
$\theta$  & 0.3 & 0.4 & 0.5 & 0.6 & 0.7 \\ \hline
NME      & 4.25  & 4.22  & \textbf{4.21}  & 4.26   & 4.23 \\ \hline
\end{tabular}}
\vspace*{-0.1in}
\caption{Evaluation on different values of $\theta$.}
\label{table:different_theta}
\vspace*{-0.15in}
\end{table}
%-------------------------------------------------------------------------
\subsubsection{Evaluation of different modules}
Evaluation on the effectiveness of different modules is shown in Table~\ref{table:Ablation}. The dataset used for ablation study is WFLW. During training and testing, faces are cropped from ground truth landmarks. Note the baseline model (model trained with MSE) underperforms the state-of-the-art. To compare with a naive weight mask without focus on hard negative pixels, we introduced a baseline weight map $WM\textsubscript{base} = \hat{H}W+1$, where $W = 10$. The major contribution comes from Adaptive Wing loss, which improves the benchmark by 0.74\%. All other modules contributed incrementally to the localization performance, our Weighted Loss Map improves 0.25\%, boundary prediction and coordinates encoding are able to contribute another 0.09\%. Our Weighted Loss Map also outperforms $WM\textsubscript{base}$ by a considerable margin, thanks to its ability to focus on hard background pixels.
\begin{table}[]
\centering
\resizebox{0.6\linewidth}{!}{%
\begin{tabular}{|c|c|}
\hline
Method       & Mean Error(\%) \\ \hline
MSE          & 5.39               \\
MSE+WM       & 5.04               \\
AW           & 4.65               \\
AW+WM\textsubscript{base}  & 4.49               \\
AW+WM        & 4.30               \\
AW+WM+B      & 4.28               \\
AW+WM+B+C    & 4.26               \\
AW+WM+B+C+CB & 4.21               \\ \hline
\end{tabular}}
\vspace*{-0.1in}
\caption{Ablation study on different methods, where AW is the Adaptive Wing Loss, WM\textsubscript{base} is the baseline weight mask, WM is our Weighted Loss Map, B is boundary integration, C is CoordConv and CB is CoordConv with boundary coordinates.}
\label{table:Ablation}
\vspace*{-0.15in}
\end{table}
%-------------------------------------------------------------------------
\subsection{Evaluation on human pose estimation}
Although this paper mainly deals with face alignment, we have also performed experiments to prove the ability of the proposed Adaptive Wing loss in another heatmap regression task, human pose estimation. We choose LSP~\cite{lsp_dataset} (using person-centric (PC) annotations) as evaluation dataset. LSP dataset consists of 11,000 training images and 1,000 testing images. Each image is labeled with 14 keypoints. The goal of this experiment is to examine the capability of the proposed Adaptive Wing loss to handle the pose estimation task compared with baseline MSE loss, rather than achieving the state-of-the-art in human pose estimation. Some other works~\cite{chu2017multi, wei2016convolutional, insafutdinov2016deepercut, pishchulin2016deepcut} obtain better results by  adding MPII~\cite{andriluka20142d} into training or as pre-training, or use re-annotated labels with high resolution images in \cite{pishchulin2016deepcut}. Besides the MSE loss baseline, we also reported baselines from methods that trained solely on the LSP dataset. We trained our model from scratch with original labeling and low resolution images to see how well our Adaptive Wing loss could handle labeling noise and low quality images. Percentage Correct Keypoints (PCK)~\cite{yang2013articulated_pck} is used as the evaluation metric with torso dimension as the normalization factor. Please refer to the supplemental materials for more implementation details. Results are shown in Table~\ref{table:LSP}. Our proposed Adaptive Wing loss significantly boosts performance compared with MSE, which proves the general applicability of the proposed Adaptive Wing loss on more heatmap regression tasks.
\begin{table}[]
\vspace{-0.05in}
\centering
\resizebox{1.0\linewidth}{!}{%
\begin{tabular}{|ccccccccc|}
\hline
Method             & Head  & Sho.  & Elb.  & Wri.  & Hip  & Knee & Ank. & Mean \\ \hline
DeepCut~\cite{pishchulin2016deepcut} & 94.6   & 86.8   & 79.9   & 75.4   & 83.5   & 82.8   & 77.9   & 83.0   \\
Pishchulin et al.~\cite{pishchulin2016deepcut} & -   & -   & -   & -   & -   & -   & -   & 84.3   \\  \hline
4HG+MSE            & 94.3    & 85.9   & 78.2   & 72.0   & 84.8   & 83.1   & 80.6   & 81.8   \\
\textbf{4HG+AW} & \textbf{96.3} & \textbf{88.7} & \textbf{81.1} & \textbf{78.2} & \textbf{88.3} & \textbf{88.1} & \textbf{86.4} & \textbf{85.9} \\\hline

\end{tabular}}
\caption{Evaluation on LSP dataset with PCK@0.2.}
\label{table:LSP}
\vspace{-0.2in}
\end{table}
%-------------------------------------------------------------------------
\section{Conclusion}
In this paper, we located two issues in the MSE loss function in heatmap regression. To resolve these issues, we proposed the Adaptive Wing loss and Weighted Loss Map for accurate localization of facial landmarks. To further improve localization results, we also introduced boundary prediction and CoordConv with boundary coordinates into our model. Experiments show that our approach is able to outperform the state-of-the-art on multiple datasets by a significant margin, using various evaluation metrics, especially on failure rate and AUC, which indicates our approach is more robust to difficult scenarios. 

\section{Acknowledgement}
This paper is partially supported by the National Science Foundation under award 1751402.

{\small
\bibliographystyle{ieee_fullname}
\bibliography{egbib}

\begin{thebibliography}{10}\itemsep=-1pt

\bibitem{andriluka20142d}
Mykhaylo Andriluka, Leonid Pishchulin, Peter Gehler, and Bernt Schiele.
\newblock 2d human pose estimation: New benchmark and state of the art
  analysis.
\newblock In {\em Proceedings of the IEEE Conference on computer Vision and
  Pattern Recognition}, pages 3686--3693, 2014.

\bibitem{LFPW}
Peter~N Belhumeur, David~W Jacobs, David~J Kriegman, and Neeraj Kumar.
\newblock Localizing parts of faces using a consensus of exemplars.
\newblock {\em IEEE transactions on pattern analysis and machine intelligence},
  35(12):2930--2940, 2013.

\bibitem{black1996robust}
Michael~J Black and Paul Anandan.
\newblock The robust estimation of multiple motions: Parametric and
  piecewise-smooth flow fields.
\newblock {\em Computer vision and image understanding}, 63(1):75--104, 1996.

\bibitem{black1996unification}
Michael~J Black and Anand Rangarajan.
\newblock On the unification of line processes, outlier rejection, and robust
  statistics with applications in early vision.
\newblock {\em International Journal of Computer Vision}, 19(1):57--91, 1996.

\bibitem{heatmap_regre_1}
Adrian Bulat and Georgios Tzimiropoulos.
\newblock Two-stage convolutional part heatmap regression for the 1st 3d face
  alignment in the wild (3dfaw) challenge.
\newblock In {\em European Conference on Computer Vision}, pages 616--624.
  Springer, 2016.

\bibitem{bulat2017binarized}
Adrian Bulat and Georgios Tzimiropoulos.
\newblock Binarized convolutional landmark localizers for human pose estimation
  and face alignment with limited resources.
\newblock In {\em The IEEE International Conference on Computer Vision (ICCV)},
  volume~1, page~4, 2017.

\bibitem{bulat2017far}
Adrian Bulat and Georgios Tzimiropoulos.
\newblock How far are we from solving the 2d \& 3d face alignment problem?(and
  a dataset of 230,000 3d facial landmarks).
\newblock In {\em International Conference on Computer Vision}, volume~1,
  page~4, 2017.

\bibitem{burgos2013robust}
Xavier~P Burgos-Artizzu, Pietro Perona, and Piotr Doll{\'a}r.
\newblock Robust face landmark estimation under occlusion.
\newblock In {\em Proceedings of the IEEE International Conference on Computer
  Vision}, pages 1513--1520, 2013.

\bibitem{cao2014face}
Xudong Cao, Yichen Wei, Fang Wen, and Jian Sun.
\newblock Face alignment by explicit shape regression.
\newblock {\em International Journal of Computer Vision}, 107(2):177--190,
  2014.

\bibitem{chu2017multi}
Xiao Chu, Wei Yang, Wanli Ouyang, Cheng Ma, Alan~L Yuille, and Xiaogang Wang.
\newblock Multi-context attention for human pose estimation.
\newblock In {\em Proceedings of the IEEE Conference on Computer Vision and
  Pattern Recognition}, pages 1831--1840, 2017.

\bibitem{face_rec_5}
Jiankang Deng, Jia Guo, and Stefanos Zafeiriou.
\newblock Arcface: Additive angular margin loss for deep face recognition.
\newblock {\em arXiv:1801.07698}, 2018.

\bibitem{deng2016m3}
Jiankang Deng, Qingshan Liu, Jing Yang, and Dacheng Tao.
\newblock M3 csr: Multi-view, multi-scale and multi-component cascade shape
  regression.
\newblock {\em Image and Vision Computing}, 47:19--26, 2016.

\bibitem{deng2017joint}
Jiankang Deng, George Trigeorgis, Yuxiang Zhou, and Stefanos Zafeiriou.
\newblock Joint multi-view face alignment in the wild.
\newblock {\em arXiv preprint arXiv:1708.06023}, 2017.

\bibitem{dong2018style}
Xuanyi Dong, Yan Yan, Wanli Ouyang, and Yi Yang.
\newblock Style aggregated network for facial landmark detection.
\newblock In {\em CVPR}, volume~2, page~6, 2018.

\bibitem{dong2018supervision}
Xuanyi Dong, Shoou-I Yu, Xinshuo Weng, Shih-En Wei, Yi Yang, and Yaser Sheikh.
\newblock Supervision-by-registration: An unsupervised approach to improve the
  precision of facial landmark detectors.

\bibitem{face_recons_1}
Pengfei Dou, Shishir~K Shah, and Ioannis~A Kakadiaris.
\newblock End-to-end 3d face reconstruction with deep neural networks.
\newblock In {\em Proc. IEEE Conference on Computer Vision and Pattern
  Recognition}, pages 21--26, 2017.

\bibitem{fan2016approaching}
Haoqiang Fan and Erjin Zhou.
\newblock Approaching human level facial landmark localization by deep
  learning.
\newblock {\em Image and Vision Computing}, 47:27--35, 2016.

\bibitem{Feng_Wing_Loss}
Zhen-Hua Feng, Josef Kittler, Muhammad Awais, Patrik Huber, and Xiao-Jun Wu.
\newblock Wing loss for robust facial landmark localisation with convolutional
  neural networks.
\newblock In {\em The IEEE Conference on Computer Vision and Pattern
  Recognition (CVPR)}, June 2018.

\bibitem{feng2017dynamic}
Zhen-Hua Feng, Josef Kittler, William Christmas, Patrik Huber, and Xiao-Jun Wu.
\newblock Dynamic attention-controlled cascaded shape regression exploiting
  training data augmentation and fuzzy-set sample weighting.
\newblock In {\em 2017 IEEE Conference on Computer Vision and Pattern
  Recognition (CVPR)}, pages 3681--3690. IEEE, 2017.

\bibitem{ganan1985bayesian}
Stuart Ganan and D McClure.
\newblock Bayesian image analysis: An application to single photon emission
  tomography.
\newblock {\em Amer. Statist. Assoc}, pages 12--18, 1985.

\bibitem{face_recons_4}
Pablo Garrido, Michael Zollh{\"o}fer, Dan Casas, Levi Valgaerts, Kiran
  Varanasi, Patrick P{\'e}rez, and Christian Theobalt.
\newblock Reconstruction of personalized 3d face rigs from monocular video.
\newblock {\em ACM Transactions on Graphics (TOG)}, 35(3):28, 2016.

\bibitem{guler2017densereg}
Riza~Alp G{\"u}ler, George Trigeorgis, Epameinondas Antonakos, Patrick Snape,
  Stefanos Zafeiriou, and Iasonas Kokkinos.
\newblock Densereg: Fully convolutional dense shape regression in-the-wild.
\newblock In {\em CVPR}, volume~2, page~5, 2017.

\bibitem{hampel2011robust}
Frank~R Hampel, Elvezio~M Ronchetti, Peter~J Rousseeuw, and Werner~A Stahel.
\newblock {\em Robust statistics: the approach based on influence functions},
  volume 196.
\newblock John Wiley \& Sons, 2011.

\bibitem{face_frontal_1}
Tal Hassner, Shai Harel, Eran Paz, and Roee Enbar.
\newblock Effective face frontalization in unconstrained images.
\newblock In {\em Proceedings of the IEEE Conference on Computer Vision and
  Pattern Recognition}, pages 4295--4304, 2015.

\bibitem{resnet}
Kaiming He, Xiangyu Zhang, Shaoqing Ren, and Jian Sun.
\newblock Deep residual learning for image recognition.
\newblock In {\em Proceedings of the IEEE conference on computer vision and
  pattern recognition}, pages 770--778, 2016.

\bibitem{honari2018improving}
Sina Honari, Pavlo Molchanov, Stephen Tyree, Pascal Vincent, Christopher Pal,
  and Jan Kautz.
\newblock Improving landmark localization with semi-supervised learning.
\newblock In {\em The IEEE Conference on Computer Vision and Pattern
  Recognition (CVPR)}, 2018.

\bibitem{insafutdinov2016deepercut}
Eldar Insafutdinov, Leonid Pishchulin, Bjoern Andres, Mykhaylo Andriluka, and
  Bernt Schiele.
\newblock Deepercut: A deeper, stronger, and faster multi-person pose
  estimation model.
\newblock In {\em European Conference on Computer Vision}, pages 34--50.
  Springer, 2016.

\bibitem{lsp_dataset}
Sam Johnson and Mark Everingham.
\newblock Clustered pose and nonlinear appearance models for human pose
  estimation.
\newblock In {\em BMVC}, volume~2, page~5, 2010.

\bibitem{pose-invariant-face-alignment-with-a-single-cnn}
Amin Jourabloo, Xiaoming Liu, Mao Ye, and Liu Ren.
\newblock Pose-invariant face alignment with a single cnn.
\newblock In {\em In Proceeding of International Conference on Computer
  Vision}, Venice, Italy, October 2017.

\bibitem{face_frontal_3}
Sanghoon Kang, Jinmook Lee, Kyeongryeol Bong, Changhyeon Kim, Youchang Kim, and
  Hoi-Jun Yoo.
\newblock Low-power scalable 3-d face frontalization processor for cnn-based
  face recognition in mobile devices.
\newblock {\em IEEE Journal on Emerging and Selected Topics in Circuits and
  Systems}, 2018.

\bibitem{kazemi2014one}
Vahid Kazemi and Josephine Sullivan.
\newblock One millisecond face alignment with an ensemble of regression trees.
\newblock In {\em Proceedings of the IEEE Conference on Computer Vision and
  Pattern Recognition}, pages 1867--1874, 2014.

\bibitem{heatmap_regre_2}
Marek Kowalski, Jacek Naruniec, and Tomasz Trzcinski.
\newblock Deep alignment network: A convolutional neural network for robust
  face alignment.
\newblock In {\em Proceedings of the International Conference on Computer
  Vision \& Pattern Recognition (CVPRW), Faces-in-the-wild Workshop/Challenge},
  volume~3, page~6, 2017.

\bibitem{kumardisentangling}
Amit Kumar and Rama Chellappa.
\newblock Disentangling 3d pose in a dendritic cnn for unconstrained 2d face
  alignment.

\bibitem{kumar2018disentangling}
Amit Kumar and Rama Chellappa.
\newblock Disentangling 3d pose in a dendritic cnn for unconstrained 2d face
  alignment.
\newblock In {\em Proceedings of the IEEE Conference on Computer Vision and
  Pattern Recognition}, pages 430--439, 2018.

\bibitem{HELEN}
Vuong Le, Jonathan Brandt, Zhe Lin, Lubomir Bourdev, and Thomas~S Huang.
\newblock Interactive facial feature localization.
\newblock In {\em European conference on computer vision}, pages 679--692.
  Springer, 2012.

\bibitem{lee2015face}
Donghoon Lee, Hyunsin Park, and Chang~D Yoo.
\newblock Face alignment using cascade gaussian process regression trees.
\newblock In {\em Proceedings of the IEEE Conference on Computer Vision and
  Pattern Recognition}, pages 4204--4212, 2015.

\bibitem{face_recons_3}
Feng Liu, Dan Zeng, Qijun Zhao, and Xiaoming Liu.
\newblock Joint face alignment and 3d face reconstruction.
\newblock In {\em European Conference on Computer Vision}, pages 545--560.
  Springer, 2016.

\bibitem{CoordConv}
Rosanne Liu, Joel Lehman, Piero Molino, Felipe~Petroski Such, Eric Frank, Alex
  Sergeev, and Jason Yosinski.
\newblock An intriguing failing of convolutional neural networks and the
  coordconv solution.
\newblock {\em arXiv preprint arXiv:1807.03247}, 2018.

\bibitem{face_rec_3}
Weiyang Liu, Yandong Wen, Zhiding Yu, Ming Li, Bhiksha Raj, and Le Song.
\newblock Sphereface: Deep hypersphere embedding for face recognition.
\newblock In {\em The IEEE Conference on Computer Vision and Pattern
  Recognition (CVPR)}, volume~1, page~1, 2017.

\bibitem{dense-face-alignment}
Yaojie Liu, Amin Jourabloo, William Ren, and Xiaoming Liu.
\newblock Dense face alignment.
\newblock In {\em In Proceeding of International Conference on Computer Vision
  Workshops}, Venice, Italy, October 2017.

\bibitem{cnn_direct_3}
Jiang-Jing Lv, Xiaohu Shao, Junliang Xing, Cheng Cheng, Xi Zhou, et~al.
\newblock A deep regression architecture with two-stage re-initialization for
  high performance facial landmark detection.
\newblock In {\em CVPR}, volume~1, page~4, 2017.

\bibitem{AFLW}
Peter M.~Roth Martin~Koestinger, Paul~Wohlhart and Horst Bischof.
\newblock {Annotated Facial Landmarks in the Wild: A Large-scale, Real-world
  Database for Facial Landmark Localization}.
\newblock In {\em {Proc. First IEEE International Workshop on Benchmarking
  Facial Image Analysis Technologies}}, 2011.

\bibitem{face_rec_2}
Iacopo Masi, Stephen Rawls, G{\'e}rard Medioni, and Prem Natarajan.
\newblock Pose-aware face recognition in the wild.
\newblock In {\em Proceedings of the IEEE Conference on Computer Vision and
  Pattern Recognition}, pages 4838--4846, 2016.

\bibitem{merget2018robust}
Daniel Merget, Matthias Rock, and Gerhard Rigoll.
\newblock Robust facial landmark detection via a fully-convolutional
  local-global context network.
\newblock In {\em Proceedings of the IEEE Conference on Computer Vision and
  Pattern Recognition}, pages 781--790, 2018.

\bibitem{XM2VTSDB}
Kieron Messer, Jiri Matas, Josef Kittler, Juergen Luettin, and Gilbert Maitre.
\newblock Xm2vtsdb: The extended m2vts database.
\newblock In {\em Second international conference on audio and video-based
  biometric person authentication}, volume 964, pages 965--966, 1999.

\bibitem{miao2018direct}
Xin Miao, Xiantong Zhen, Xianglong Liu, Cheng Deng, Vassilis Athitsos, and Heng
  Huang.
\newblock Direct shape regression networks for end-to-end face alignment.
\newblock In {\em Proceedings of the IEEE Conference on Computer Vision and
  Pattern Recognition}, pages 5040--5049, 2018.

\bibitem{Hourglass}
Alejandro Newell, Kaiyu Yang, and Jia Deng.
\newblock Stacked hourglass networks for human pose estimation.
\newblock In {\em European Conference on Computer Vision}, pages 483--499.
  Springer, 2016.

\bibitem{pishchulin2016deepcut}
Leonid Pishchulin, Eldar Insafutdinov, Siyu Tang, Bjoern Andres, Mykhaylo
  Andriluka, Peter~V Gehler, and Bernt Schiele.
\newblock Deepcut: Joint subset partition and labeling for multi person pose
  estimation.
\newblock In {\em Proceedings of the IEEE Conference on Computer Vision and
  Pattern Recognition}, pages 4929--4937, 2016.

\bibitem{ren2014face}
Shaoqing Ren, Xudong Cao, Yichen Wei, and Jian Sun.
\newblock Face alignment at 3000 fps via regressing local binary features.
\newblock In {\em Proceedings of the IEEE Conference on Computer Vision and
  Pattern Recognition}, pages 1685--1692, 2014.

\bibitem{300W_inter_pupil}
Shaoqing Ren, Xudong Cao, Yichen Wei, and Jian Sun.
\newblock Face alignment via regressing local binary features.
\newblock {\em IEEE Transactions on Image Processing}, 25(3):1233--1245, 2016.

\bibitem{300W_partition}
Shaoqing Ren, Xudong Cao, Yichen Wei, and Jian Sun.
\newblock Face alignment via regressing local binary features.
\newblock {\em IEEE Transactions on Image Processing}, 25(3):1233--1245, 2016.

\bibitem{face_recons_2}
Joseph Roth, Yiying Tong, and Xiaoming Liu.
\newblock Unconstrained 3d face reconstruction.
\newblock In {\em The IEEE Conference on Computer Vision and Pattern
  Recognition (CVPR)}, June 2015.

\bibitem{300W}
Christos Sagonas, Georgios Tzimiropoulos, Stefanos Zafeiriou, and Maja Pantic.
\newblock 300 faces in-the-wild challenge: The first facial landmark
  localization challenge.
\newblock In {\em 2013 IEEE International Conference on Computer Vision
  Workshops}, pages 397--403. IEEE, 2013.

\bibitem{heatmap_regre_4}
Xiaohu Shao, Junliang Xing, Jiang-Jing Lv, Chunlin Xiao, Pengcheng Liu, Youji
  Feng, Cheng Cheng, and F Si.
\newblock Unconstrained face alignment without face detection.
\newblock In {\em CVPR Workshops}, pages 2069--2077, 2017.

\bibitem{face_rec_1}
Yaniv Taigman, Ming Yang, Marc'Aurelio Ranzato, and Lior Wolf.
\newblock Deepface: Closing the gap to human-level performance in face
  verification.
\newblock In {\em Proceedings of the IEEE conference on computer vision and
  pattern recognition}, pages 1701--1708, 2014.

\bibitem{tang2018quantized}
Zhiqiang Tang, Xi Peng, Shijie Geng, Lingfei Wu, Shaoting Zhang, and Dimitris
  Metaxas.
\newblock Quantized densely connected u-nets for efficient landmark
  localization.
\newblock In {\em European Conference on Computer Vision (ECCV)}, 2018.

\bibitem{RMSProp}
Tijmen Tieleman and Geoffrey Hinton.
\newblock Lecture 6.5-rmsprop: Divide the gradient by a running average of its
  recent magnitude.
\newblock {\em COURSERA: Neural networks for machine learning}, 4(2):26--31,
  2012.

\bibitem{trigeorgis2016mnemonic}
George Trigeorgis, Patrick Snape, Mihalis~A Nicolaou, Epameinondas Antonakos,
  and Stefanos Zafeiriou.
\newblock Mnemonic descent method: A recurrent process applied for end-to-end
  face alignment.
\newblock In {\em Proceedings of the IEEE Conference on Computer Vision and
  Pattern Recognition}, pages 4177--4187, 2016.

\bibitem{valle2018deeply}
Roberto Valle and M Jos{\'e}.
\newblock A deeply-initialized coarse-to-fine ensemble of regression trees for
  face alignment.
\newblock In {\em Proceedings of the European Conference on Computer Vision
  (ECCV)}, pages 585--601, 2018.

\bibitem{face_frontal_2}
Yiming Wang, Hui Yu, Junyu Dong, Brett Stevens, and Honghai Liu.
\newblock Facial expression-aware face frontalization.
\newblock In {\em Asian Conference on Computer Vision}, pages 375--388.
  Springer, 2016.

\bibitem{wei2016convolutional}
Shih-En Wei, Varun Ramakrishna, Takeo Kanade, and Yaser Sheikh.
\newblock Convolutional pose machines.
\newblock In {\em Proceedings of the IEEE Conference on Computer Vision and
  Pattern Recognition}, pages 4724--4732, 2016.

\bibitem{LAB}
Wayne Wu, Chen Qian, Shuo Yang, Quan Wang, Yici Cai, and Qiang Zhou.
\newblock Look at boundary: A boundary-aware face alignment algorithm.
\newblock In {\em The IEEE Conference on Computer Vision and Pattern
  Recognition (CVPR)}, June 2018.

\bibitem{wu2017leveraging}
Wenyan Wu and Shuo Yang.
\newblock Leveraging intra and inter-dataset variations for robust face
  alignment.
\newblock In {\em Proceedings of the International Conference on Computer
  Vision \& Pattern Recognition (CVPR), Faces-in-the-wild Workshop/Challenge},
  volume~3, page~6, 2017.

\bibitem{wu2015robust}
Yue Wu and Qiang Ji.
\newblock Robust facial landmark detection under significant head poses and
  occlusion.
\newblock In {\em Proceedings of the IEEE International Conference on Computer
  Vision}, pages 3658--3666, 2015.

\bibitem{GoDP}
Yuhang Wu, Shishir~K Shah, and Ioannis~A Kakadiaris.
\newblock Godp: Globally optimized dual pathway deep network architecture for
  facial landmark localization in-the-wild.
\newblock {\em Image and Vision Computing}, 73:1--16, 2018.

\bibitem{xiao2016robust}
Shengtao Xiao, Jiashi Feng, Junliang Xing, Hanjiang Lai, Shuicheng Yan, and
  Ashraf Kassim.
\newblock Robust facial landmark detection via recurrent attentive-refinement
  networks.
\newblock In {\em European conference on computer vision}, pages 57--72.
  Springer, 2016.

\bibitem{xiong2013supervised}
Xuehan Xiong and Fernando De~la Torre.
\newblock Supervised descent method and its applications to face alignment.
\newblock In {\em Proceedings of the IEEE conference on computer vision and
  pattern recognition}, pages 532--539, 2013.

\bibitem{heatmap_regre_3}
Jing Yang, Qingshan Liu, and Kaihua Zhang.
\newblock Stacked hourglass network for robust facial landmark localisation.
\newblock In {\em Computer Vision and Pattern Recognition Workshops (CVPRW),
  2017 IEEE Conference on}, pages 2025--2033. IEEE, 2017.

\bibitem{yang2017stacked}
Jing Yang, Qingshan Liu, and Kaihua Zhang.
\newblock Stacked hourglass network for robust facial landmark localisation.
\newblock In {\em Computer Vision and Pattern Recognition Workshops (CVPRW),
  2017 IEEE Conference on}, pages 2025--2033. IEEE, 2017.

\bibitem{face_rec_4}
Jiaolong Yang, Peiran Ren, Dongqing Zhang, Dong Chen, Fang Wen, Hongdong Li,
  and Gang Hua.
\newblock Neural aggregation network for video face recognition.
\newblock In {\em CVPR}, volume~4, page~7, 2017.

\bibitem{yang2013articulated_pck}
Yi Yang and Deva Ramanan.
\newblock Articulated human detection with flexible mixtures of parts.
\newblock {\em IEEE transactions on pattern analysis and machine intelligence},
  35(12):2878--2890, 2013.

\bibitem{zhang2014coarse}
Jie Zhang, Shiguang Shan, Meina Kan, and Xilin Chen.
\newblock Coarse-to-fine auto-encoder networks (cfan) for real-time face
  alignment.
\newblock In {\em European Conference on Computer Vision}, pages 1--16.
  Springer, 2014.

\bibitem{zhang2014facial}
Zhanpeng Zhang, Ping Luo, Chen~Change Loy, and Xiaoou Tang.
\newblock Facial landmark detection by deep multi-task learning.
\newblock In {\em European Conference on Computer Vision}, pages 94--108.
  Springer, 2014.

\bibitem{zhang2016learning}
Zhanpeng Zhang, Ping Luo, Chen~Change Loy, and Xiaoou Tang.
\newblock Learning deep representation for face alignment with auxiliary
  attributes.
\newblock {\em IEEE transactions on pattern analysis and machine intelligence},
  38(5):918--930, 2016.

\bibitem{zhu2015face}
Shizhan Zhu, Cheng Li, Chen Change~Loy, and Xiaoou Tang.
\newblock Face alignment by coarse-to-fine shape searching.
\newblock In {\em Proceedings of the IEEE Conference on Computer Vision and
  Pattern Recognition}, pages 4998--5006, 2015.

\bibitem{Unconstrained_AFLW}
Shizhan Zhu, Cheng Li, Chen-Change Loy, and Xiaoou Tang.
\newblock Unconstrained face alignment via cascaded compositional learning.
\newblock In {\em Proceedings of the IEEE Conference on Computer Vision and
  Pattern Recognition}, pages 3409--3417, 2016.

\bibitem{zhu2016unconstrained}
Shizhan Zhu, Cheng Li, Chen-Change Loy, and Xiaoou Tang.
\newblock Unconstrained face alignment via cascaded compositional learning.
\newblock In {\em Proceedings of the IEEE Conference on Computer Vision and
  Pattern Recognition}, pages 3409--3417, 2016.

\bibitem{AFW}
Xiangxin Zhu and Deva Ramanan.
\newblock Face detection, pose estimation, and landmark localization in the
  wild.
\newblock In {\em Computer Vision and Pattern Recognition (CVPR), 2012 IEEE
  Conference on}, pages 2879--2886. IEEE, 2012.

\end{thebibliography}
}

\section{Supplementary Material}
\subsection{Implementation Detail of CoordConv on Boundary Information}
\label{sec:coordconv_boundary}
\begin{figure}[!htb]
    \centering
    \includegraphics[width=1.0\linewidth]{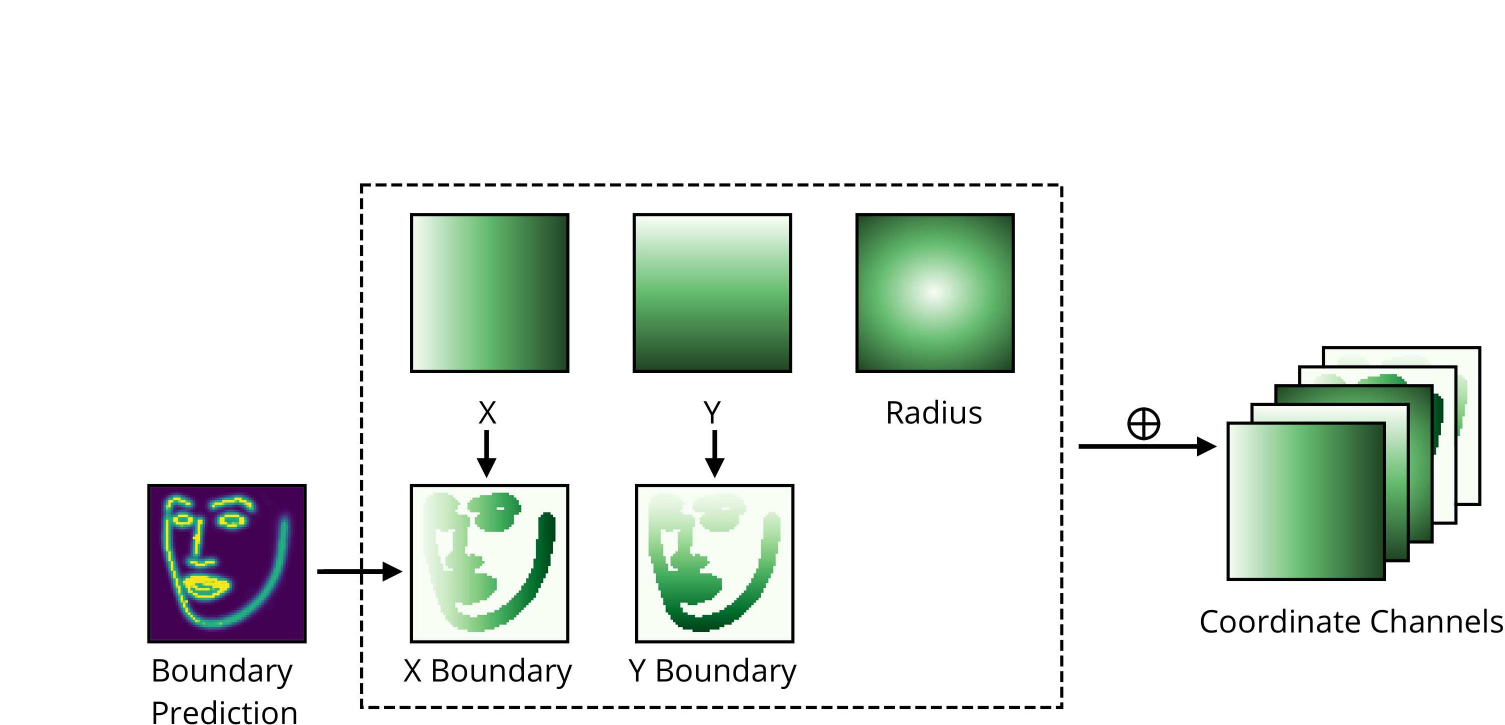}
    \caption{\textbf{CoodConv with Boundary Information}. $X$ Boundary and $Y$ Boundary are generated from $X$ coordinate channel and $Y$ coordinate channel respectively by a binary mask created from boundary prediction from the previous Hourglass module. The mask is generated by thresholding boundary prediction with a value of 0.05.  (Best viewed in color).}
    \label{fig:coordconv}
\end{figure}

In addition to original CoordConv~\cite{CoordConv}, we add two coordinate encoding channels with boundary information. A visualization of this process is shown in Figure~\ref{fig:coordconv}

\subsection{Datasets Used in Our Experiments}
\label{sec:face_dataset}
The \textbf{COFW}~\cite{burgos2013robust} dataset includes 1,345 training images and 507 testing images annotated with 29 landmarks. This dataset is aimed to test the effectiveness of face alignment algorithms on faces with large pose and heavy occlusion. Various types of occlusions are introduced and result in a 23\% occlusion on facial parts in average.

The \textbf{300W~\cite{300W}} is widely used as a 2D face alignment benchmark with 68 annotated landmarks. 300W consists of the following subsets: LFPW~\cite{LFPW}, HELEN~\cite{HELEN}, AFW~\cite{AFW}, XM2VTS~\cite{XM2VTSDB} and an additional dataset with 135 images with large pose, occlusion and expressions called iBUG. To compare with other approaches, we adopt the widely used protocol described in~\cite{300W_partition} to train and evaluate our approach. More specifically, we use the training dataset of LFPW, HELEN, and the full AFW dataset as training dataset, and the test dataset of LFPW, HELEN and the full iBUG dataset as full test dataset. The full test dataset is then further split into two subsets, the test dataset of LFPW and HELEN is called the common test dataset, and iBUG is called the challenge test dataset. There is also a 300W private test dataset for the 300W contest, which contains 300 indoor and 300 outdoor faces. We also evaluated our approach on this dataset.

The \textbf{WFLW~\cite{LAB}} is a newly introduced dataset with 98 manually annotated landmarks that constitutes of 7,500 training images and 2,500 testing images. In addition to denser annotations, it also provides attribute annotations including pose, expression, illumination, make-up, occlusion and blur. The six different subsets can be used for analyzing algorithm performance on subsets with different properties separately. The WFLW is considered more difficult than commonly used datasets such as AFLW and 300W due to its more densely annotated landmarks and difficult faces with occlusion, blur, large pose, makeup, expression and illumination. 

For the \textbf{LSP~\cite{lsp_dataset}} dataset, we used original label from author's official website\footnote{\url{http://sam.johnson.io/research/lsp.html}}\footnote{\url{http://sam.johnson.io/research/lspet.html}}. Although images with original resolutions are also provided, we choose not to use them. Also, we did not use re-annotated labels on LSP extended 10,000 training images from \cite{pishchulin2016deepcut}. Note that occluded keypoints are annotated in LSP original dataset but not in LSP extended training dataset. During training, we did not calculate loss on occluded keypoints for LSP extended training dataset. During training and testing, we did not follow~\cite{peng2018jointly} to crop single person from images with multiple persons to retain the difficulties of this dataset. Data augmentations is performed similarly to training with face alignment datasets.

\subsection{Evaluation on AFLW}
\label{sec:eval_aflw}
The \textbf{AFLW~\cite{AFLW}} dataset contains 24,368 faces with large poses. All faces are annotated by up to 21 landmarks per image, while the occluded landmarks were not labeled. For fair comparison with other methods we adopt the protocol from~\cite{Unconstrained_AFLW}, which provides revised annotations with 19 landmarks. The training dataset contains 20,000 images, the full testing dataset contains 4,368 iamges. A subset of 1,314 frontal faces (no landmarks are occluded) are selected from the full test dataset as the frontal test set.

\begin{table}[H]
\centering
\resizebox{0.7\linewidth}{!}{%
\setlength\tabcolsep{1.5pt}
\begin{tabular}{|ccc|}
\hline
Method  & Full(\%) & Frontal(\%) \\ \hline
RCPR\textsubscript{CVPR 13} \cite{burgos2013robust}    &    3.73       &     2.87         \\
ERT\textsubscript{CVPR 14}~\cite{kazemi2014one}     &    4.35       &     2.75         \\
LBF\textsubscript{CVPR 14}~\cite{ren2014face}     &    4.25       &     2.74         \\
CFSS\textsubscript{CVPR 15}~\cite{zhu2015face}    &    3.92       &     2.68         \\
CCL\textsubscript{CVPR 16}~\cite{zhu2016unconstrained}     &    2.72       &     2.17         \\
TSR\textsubscript{CVPR 17}~\cite{cnn_direct_3}     &    2.17       &      -           \\
DAC-OSR\textsubscript{CVPR 17}~\cite{feng2017dynamic} &    2.27       &     1.81         \\
DCFE\textsubscript{ECCV 18}~\cite{valle2018deeply} &       2.17 & - \\
CPM+SBR\textsubscript{CVPR 18}~\cite{dong2018supervision} &  2.14     & - \\ 
SAN\textsubscript{CVPR 18}~\cite{dong2018style} & 1.91 & 1.85 \\
DSRN\textsubscript{CVPR 18}~\cite{miao2018direct} & 1.86 & - \\
LAB\textsubscript{CVPR 18}~\cite{LAB}     &    1.85       &     1.62         \\
Wing\textsubscript{CVPR 18}~\cite{Feng_Wing_Loss}    &    1.65       &      -           \\
\small{RCN\textsuperscript{+}(L+ELT+A)\textsubscript{CVPR 18}~\cite{honari2018improving}} & 1.59 & - \\ \hline
\textbf{AWing(Ours)} & \textbf{1.53} & \textbf{1.38} \\ \hline
\end{tabular}}
\caption{Mean error(\%) on the AFLW testset}
\label{table:AFLW}
\end{table}

Evaluation results on the AFLW dataset are shown in Table~\ref{table:AFLW}. For AFLW dataset, we created boundary with a different scheme compared with Wu\etal~\cite{LAB} since insufficient landmarks are provided to generate all 14 boundary lines. We only use landmarks to generate left/right eyebrow, left/right eye line and noise bottom line. Even though we only have limited boundary information from 19 landmarks, our method is able to outperform the state-of-the-art methods in a large margin, which prove the robustness of our method to faces with large poses.

\subsection{Additional Ablation Study}
\label{sec:ablation}
\subsubsection{Effectiveness of Adaptive Wing loss on Training}
Table~\ref{table:AW_training_efficiency} shows the effectiveness of our Adaptive Wing loss compare with MSE in terms of training loss w.r.t. the number of training epochs. Model trained with the Adaptive Wing loss is able to reduce the pixel-wise average MSE loss for almost 30\%, and more than 23\% on foreground pixels. Especially, this improvement comes at a mere $50$ epochs, showing that the AWing loss improves convergence speed.
\begin{table}[]
\centering
\setlength\tabcolsep{3pt}
\resizebox{1.0\linewidth}{!}{%
\begin{tabular}{|c|ccccc|}
\hline
\backslashbox{Loss}{Epoch}     & 10          & 50           & 100          & 150          & 200          \\ \hline
MSE\_all & 0.018       & 0.018        & 0.014        & 0.014        & 0.014        \\ \hline
AW\_all  & 0.018(-)    & 0.013(\textcolor{OliveGreen}{$\downarrow$27\%}) & 0.011(\textcolor{OliveGreen}{$\downarrow$21\%}) & 0.010(\textcolor{OliveGreen}{$\downarrow$28\%}) & 0.010(\textcolor{OliveGreen}{$\downarrow$28\%}) \\ \hline
MSE\_fg  & 1.17        & 1.25         & 0.95         & 0.94         & 0.92         \\ \hline
AW\_fg   & 1.13(\textcolor{OliveGreen}{$\downarrow$3\%}) & 0.87(\textcolor{OliveGreen}{$\downarrow$30\%})  & 0.74(\textcolor{OliveGreen}{$\downarrow$22\%})  & 0.72(\textcolor{OliveGreen}{$\downarrow$23\%})  & 0.71(\textcolor{OliveGreen}{$\downarrow$23\%}) \\ \hline
\end{tabular}}
\caption{Training loss comparison. For fair comparison, the losses are evaluated with MSE. Model are trained with original stacked HG without weight map. Subscript \_fg and \_all stand for foreground pixels and all pixels respectively.}
\label{table:AW_training_efficiency}
\end{table}

\subsubsection{Robustness of Adaptive Wing loss on datasets with manually added annotation noise}
We experimented our Adaptive Wing loss on the WFLW dataset with manually added labeling noise. The dataset is generated by randomly shifting $S\%$ of the inter-ocular distances from $P\%$ of the points with a random angle.
\begin{table}[htb]
\centering
\resizebox{0.75\linewidth}{!}{%
\begin{tabular}{|c|c|c|c|c|}
\hline
P(\%)/S(\%)    & 0/0 & 10/10  &20/20  &30/30 \\ \hline
%MSE         & & 4.87  & 4.99 & 5.46            \\ \hline
AWing        & 4.65  & 4.64 & 4.66 & 4.86            \\ \hline
\end{tabular}}
\caption{AWing on the WFLW dataset with noise, without Weighted Loss Map, CoordConv and boundary.}
\label{table:aw_mse_noise}
\end{table}\\
\subsubsection{Experiment on different number of HG stacks}
\begin{table*}[]
\centering
\setlength\tabcolsep{3pt}
\begin{tabular}{|c|ccc|c|c|c|c|c|}
\hline
\multirow{2}{*}{} & \multicolumn{3}{c|}{300W}         & \multirow{2}{*}{\begin{tabular}[c]{@{}c@{}}300W\\ Private\end{tabular}} & \multirow{2}{*}{WFLW} & \multirow{2}{*}{COFW} & \multirow{2}{*}{\begin{tabular}[c]{@{}c@{}}GPU\\ Runtime  (FPS)\end{tabular}} \\ \cline{2-4}
                  & Common    & Challange & Full      &                                                                         &                       &                       &           \\ \hline
Previous Best   & 3.27/2.90 & 7.18/5.15 & 4.04/3.35      & 3.88 & 5.11 & 5.27 & -  \\
AWing-1HG   & 3.89/2.81 & 6.80/4.72 & 4.46/3.18      & 3.74 & 4.50 & 5.18 & 120.47  \\
AWing-2HGs  & 3.84/2.77 & 6.61/4.58 & 4.38/3.12      & 3.61 & 4.29 & 5.08 & 63.79 \\
AWing-3HGs  & 3.79/2.73 & 6.61/4.58 & 4.34/3.10      & 3.59 & 4.24 & 5.01 & 45.29 \\
AWing-4HGs  & 3.77/2.72 & 6.52/4.52 & 4.31/3.07 & 3.56 & 4.21 & 4.94 & 34.50 \\ \hline
\end{tabular}
\caption{\textbf{NME (\%) on different number of stacks}. The NMEs of 300W are normalized by inter-pupil/inter-ocular distance, the NMEs of COFW are normalized by inter-pupil distance, and the NMEs of 300W Private and WFLW are normlaized by inter-ocular distance. NMEs in the "Previous Best" row are selected from Table 1 to 4 in our main paper. Runtime is evaluated on Nvidia GTX 1080Ti graphics card with batch size of 1.}
\label{table:ablation_stacks}
\end{table*}
We compare the performance of different number of stacks of HG module (see details in Table~\ref{table:ablation_stacks}). With reduced number of HGs, the performance of our approach remains outstanding. Even with only one HG block, our approach still outperforms previous state-of-the-arts in all datasets except the common subset and the full dataset of 300W. Note that the one HG model is able to run at 120 FPS with Nvidia GTX 1080Ti graphics card. The result reflects the effectiveness of our approach on limited computation resources.

\subsection{Result Visualization}
\label{sec:result_visualization}
\begin{figure*}[t]
\captionsetup[subfigure]{justification=centering}
\begin{subfigure}{0.167\linewidth}
\centering
\includegraphics[width=1.0\linewidth]{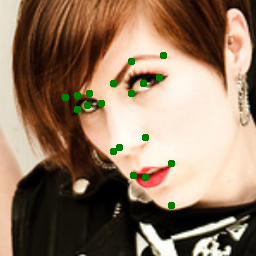}
\end{subfigure}%
\begin{subfigure}{0.167\linewidth}
\centering
\includegraphics[width=1.0\linewidth]{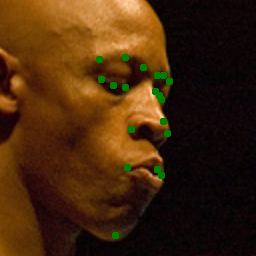}
\end{subfigure}%
\begin{subfigure}{0.167\linewidth}
\centering
\includegraphics[width=1.0\linewidth]{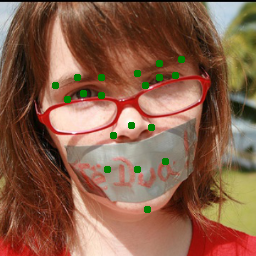}
\end{subfigure}%
\begin{subfigure}{0.167\linewidth}
\centering
\includegraphics[width=1.0\linewidth]{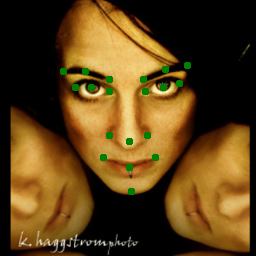}
\end{subfigure}%
\begin{subfigure}{0.167\linewidth}
\centering
\includegraphics[width=1.0\linewidth]{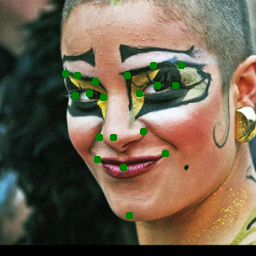}
\end{subfigure}%
\begin{subfigure}{0.167\linewidth}
\centering
\includegraphics[width=1.0\linewidth]{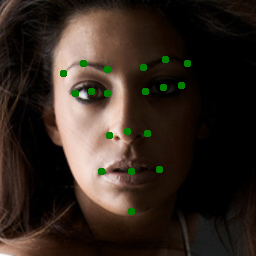}
\end{subfigure}%

\begin{subfigure}{0.167\linewidth}
\centering
\includegraphics[width=1.0\linewidth]{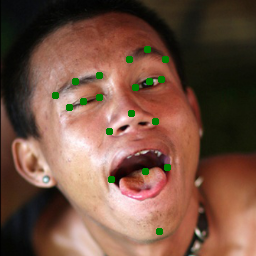}
\end{subfigure}%
\begin{subfigure}{0.167\linewidth}
\centering
\includegraphics[width=1.0\linewidth]{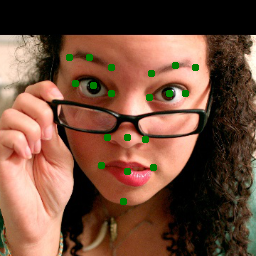}
\end{subfigure}%
\begin{subfigure}{0.167\linewidth}
\centering
\includegraphics[width=1.0\linewidth]{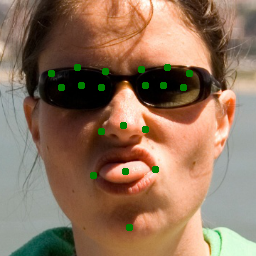}
\end{subfigure}%
\begin{subfigure}{0.167\linewidth}
\centering
\includegraphics[width=1.0\linewidth]{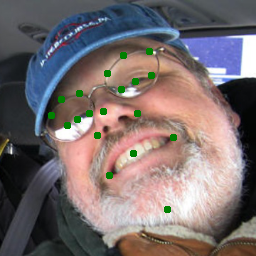}
\end{subfigure}%
\begin{subfigure}{0.167\linewidth}
\centering
\includegraphics[width=1.0\linewidth]{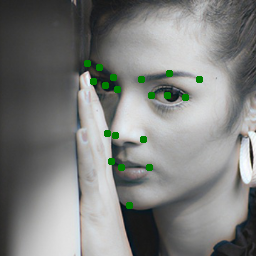}
\end{subfigure}%
\begin{subfigure}{0.167\linewidth}
\centering
\includegraphics[width=1.0\linewidth]{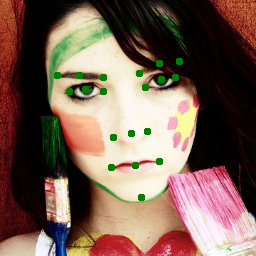}
\end{subfigure}%

\begin{subfigure}{0.167\linewidth}
\centering
\includegraphics[width=1.0\linewidth]{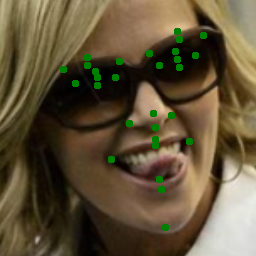}
\end{subfigure}%
\begin{subfigure}{0.167\linewidth}
\centering
\includegraphics[width=1.0\linewidth]{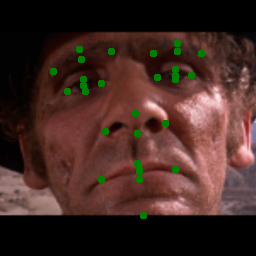}
\end{subfigure}%
\begin{subfigure}{0.167\linewidth}
\centering
\includegraphics[width=1.0\linewidth]{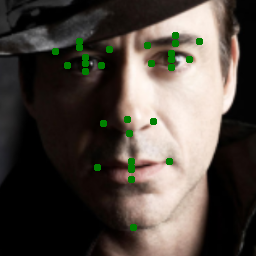}
\end{subfigure}%
\begin{subfigure}{0.167\linewidth}
\centering
\includegraphics[width=1.0\linewidth]{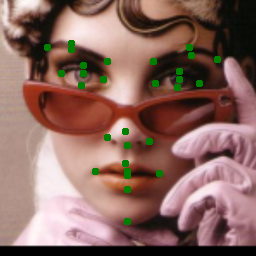}
\end{subfigure}%
\begin{subfigure}{0.167\linewidth}
\centering
\includegraphics[width=1.0\linewidth]{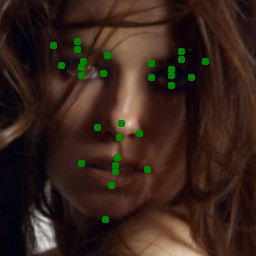}
\end{subfigure}%
\begin{subfigure}{0.167\linewidth}
\centering
\includegraphics[width=1.0\linewidth]{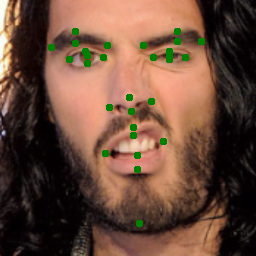}
\end{subfigure}%

\begin{subfigure}{0.167\linewidth}
\centering
\includegraphics[width=1.0\linewidth]{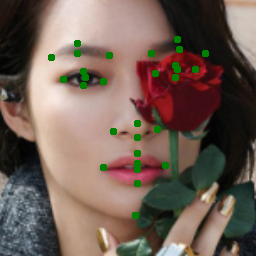}
\end{subfigure}%
\begin{subfigure}{0.167\linewidth}
\centering
\includegraphics[width=1.0\linewidth]{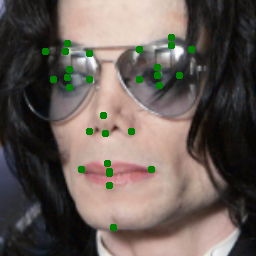}
\end{subfigure}%
\begin{subfigure}{0.167\linewidth}
\centering
\includegraphics[width=1.0\linewidth]{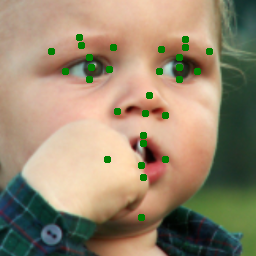}
\end{subfigure}%
\begin{subfigure}{0.167\linewidth}
\centering
\includegraphics[width=1.0\linewidth]{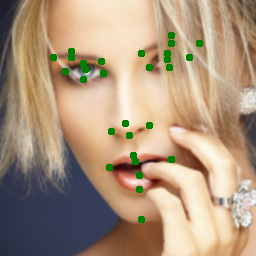}
\end{subfigure}%
\begin{subfigure}{0.167\linewidth}
\centering
\includegraphics[width=1.0\linewidth]{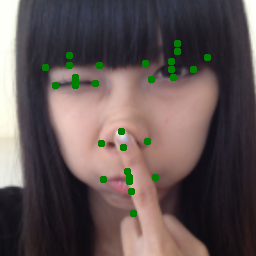}
\end{subfigure}%
\begin{subfigure}{0.167\linewidth}
\centering
\includegraphics[width=1.0\linewidth]{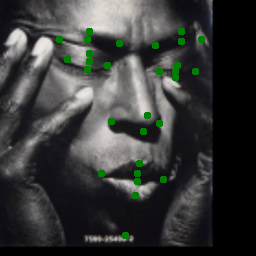}
\end{subfigure}%

\begin{subfigure}{0.167\linewidth}
\centering
\includegraphics[width=1.0\linewidth]{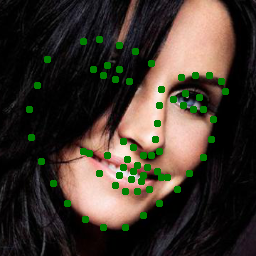}
\end{subfigure}%
\begin{subfigure}{0.167\linewidth}
\centering
\includegraphics[width=1.0\linewidth]{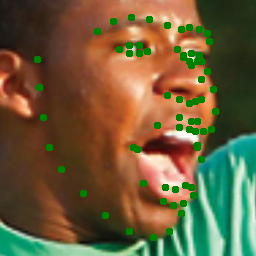}
\end{subfigure}%
\begin{subfigure}{0.167\linewidth}
\centering
\includegraphics[width=1.0\linewidth]{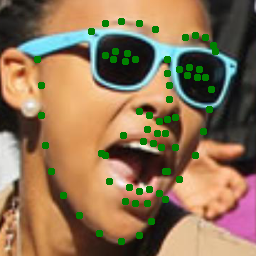}
\end{subfigure}%
\begin{subfigure}{0.167\linewidth}
\centering
\includegraphics[width=1.0\linewidth]{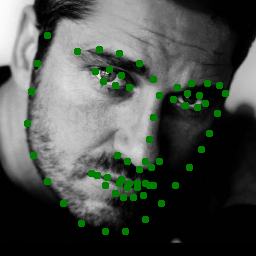}
\end{subfigure}%
\begin{subfigure}{0.167\linewidth}
\centering
\includegraphics[width=1.0\linewidth]{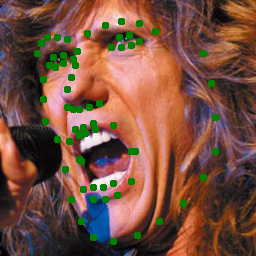}
\end{subfigure}%
\begin{subfigure}{0.167\linewidth}
\centering
\includegraphics[width=1.0\linewidth]{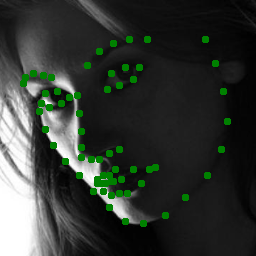}
\end{subfigure}%

\begin{subfigure}{0.167\linewidth}
\centering
\includegraphics[width=1.0\linewidth]{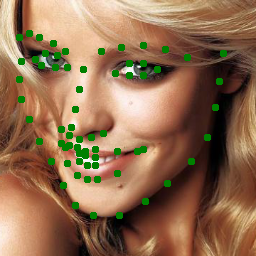}
\end{subfigure}%
\begin{subfigure}{0.167\linewidth}
\centering
\includegraphics[width=1.0\linewidth]{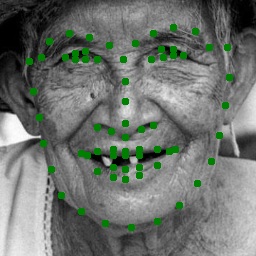}
\end{subfigure}%
\begin{subfigure}{0.167\linewidth}
\centering
\includegraphics[width=1.0\linewidth]{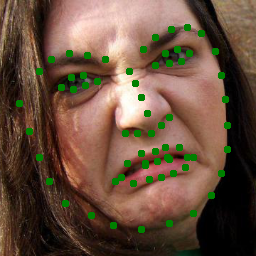}
\end{subfigure}%
\begin{subfigure}{0.167\linewidth}
\centering
\includegraphics[width=1.0\linewidth]{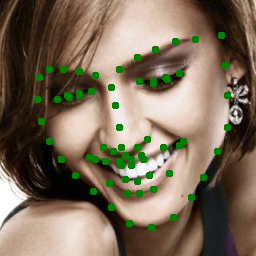}
\end{subfigure}%
\begin{subfigure}{0.167\linewidth}
\centering
\includegraphics[width=1.0\linewidth]{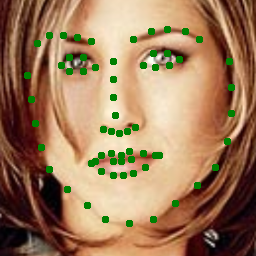}
\end{subfigure}%
\begin{subfigure}{0.167\linewidth}
\centering
\includegraphics[width=1.0\linewidth]{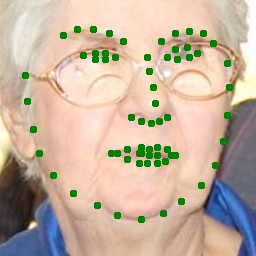}
\end{subfigure}%

\caption{\textbf{Result visualization 1}. Row 1-2: AFLW dataset, row 3-4: COFW dataset, row 5-6: 300W dataset.}
\label{fig:visualization1}
\end{figure*}

\begin{figure*}[t]
\captionsetup[subfigure]{justification=centering}
\begin{subfigure}{0.167\linewidth}
\centering
\includegraphics[width=1.0\linewidth]{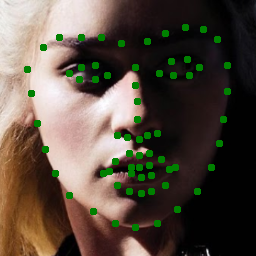}
\end{subfigure}%
\begin{subfigure}{0.167\linewidth}
\centering
\includegraphics[width=1.0\linewidth]{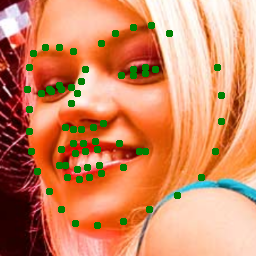}
\end{subfigure}%
\begin{subfigure}{0.167\linewidth}
\centering
\includegraphics[width=1.0\linewidth]{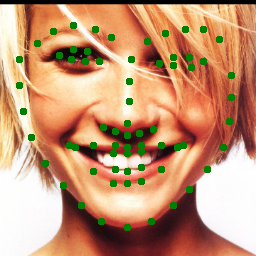}
\end{subfigure}%
\begin{subfigure}{0.167\linewidth}
\centering
\includegraphics[width=1.0\linewidth]{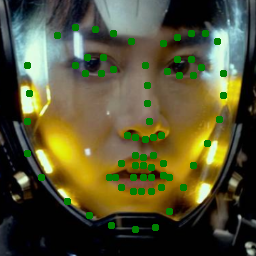}
\end{subfigure}%
\begin{subfigure}{0.167\linewidth}
\centering
\includegraphics[width=1.0\linewidth]{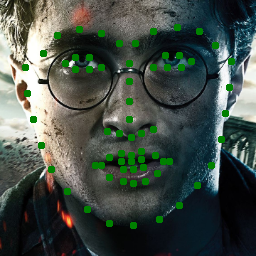}
\end{subfigure}%
\begin{subfigure}{0.167\linewidth}
\centering
\includegraphics[width=1.0\linewidth]{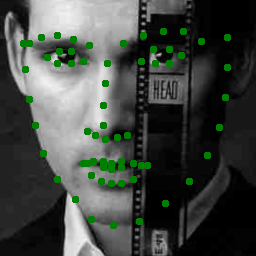}
\end{subfigure}%

\begin{subfigure}{0.167\linewidth}
\centering
\includegraphics[width=1.0\linewidth]{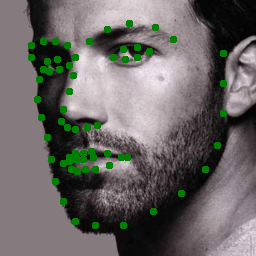}
\end{subfigure}%
\begin{subfigure}{0.167\linewidth}
\centering
\includegraphics[width=1.0\linewidth]{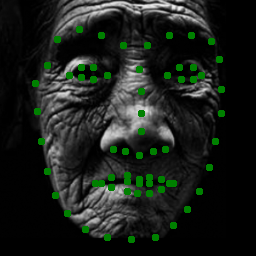}
\end{subfigure}%
\begin{subfigure}{0.167\linewidth}
\centering
\includegraphics[width=1.0\linewidth]{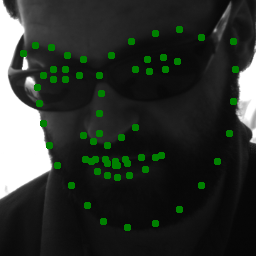}
\end{subfigure}%
\begin{subfigure}{0.167\linewidth}
\centering
\includegraphics[width=1.0\linewidth]{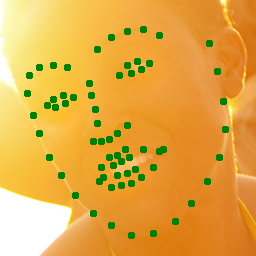}
\end{subfigure}%
\begin{subfigure}{0.167\linewidth}
\centering
\includegraphics[width=1.0\linewidth]{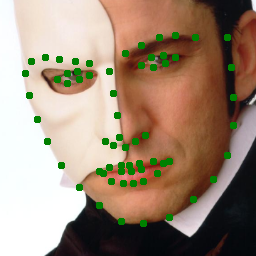}
\end{subfigure}%
\begin{subfigure}{0.167\linewidth}
\centering
\includegraphics[width=1.0\linewidth]{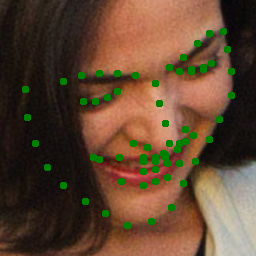}
\end{subfigure}%

\begin{subfigure}{0.167\linewidth}
\centering
\includegraphics[width=1.0\linewidth]{visual/wflw_1.png}
\end{subfigure}%
\begin{subfigure}{0.167\linewidth}
\centering
\includegraphics[width=1.0\linewidth]{visual/wflw_2.png}
\end{subfigure}%
\begin{subfigure}{0.167\linewidth}
\centering
\includegraphics[width=1.0\linewidth]{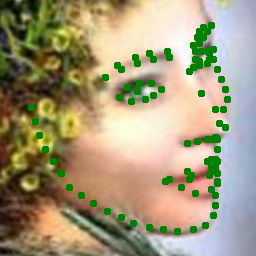}
\end{subfigure}%
\begin{subfigure}{0.167\linewidth}
\centering
\includegraphics[width=1.0\linewidth]{visual/wflw_4.png}
\end{subfigure}%
\begin{subfigure}{0.167\linewidth}
\centering
\includegraphics[width=1.0\linewidth]{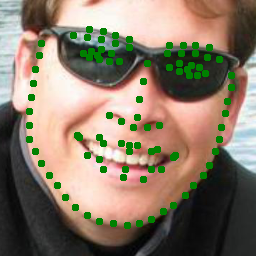}
\end{subfigure}%
\begin{subfigure}{0.167\linewidth}
\centering
\includegraphics[width=1.0\linewidth]{visual/wflw_6.png}
\end{subfigure}%

\begin{subfigure}{0.167\linewidth}
\centering
\includegraphics[width=1.0\linewidth]{visual/wflw_7.png}
\end{subfigure}%
\begin{subfigure}{0.167\linewidth}
\centering
\includegraphics[width=1.0\linewidth]{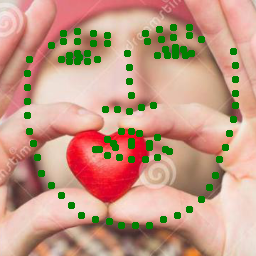}
\end{subfigure}%
\begin{subfigure}{0.167\linewidth}
\centering
\includegraphics[width=1.0\linewidth]{visual/wflw_9.png}
\end{subfigure}%
\begin{subfigure}{0.167\linewidth}
\centering
\includegraphics[width=1.0\linewidth]{visual/wflw_10.png}
\end{subfigure}%
\begin{subfigure}{0.167\linewidth}
\centering
\includegraphics[width=1.0\linewidth]{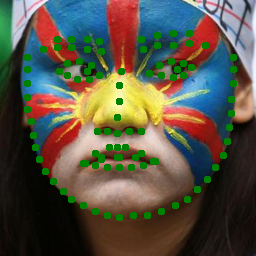}
\end{subfigure}%
\begin{subfigure}{0.167\linewidth}
\centering
\includegraphics[width=1.0\linewidth]{visual/wflw_12.png}
\end{subfigure}%
\caption{\textbf{Result visualization 2}. Row 1-2: 300W private dataset, row 3-4: WFLW dataset.}
\label{fig:visualization2}
\end{figure*}

For visualization purpose, some localization results are shown in Figure~\ref{fig:visualization1} and Figure~\ref{fig:visualization2}

\end{document}